%% file: conference.tex
\documentclass{article} 
\usepackage{conference,times}

\input{math_commands.tex}

\usepackage{hyperref}
\usepackage{url}
\usepackage{graphicx}
\usepackage{graphicx}
\usepackage{booktabs}
\usepackage{latexsym}
\usepackage{array}
\usepackage{amsmath}

\usepackage{xcolor}
\usepackage{enumitem}
\usepackage{subcaption}
\usepackage{comment}
\usepackage{verbatim}
\usepackage{booktabs}
\usepackage{array}
\usepackage{tabularx}
\usepackage{float}
\usepackage{wrapfig}
\usepackage{graphicx}
\usepackage{tcolorbox}
\newtcolorbox{takeawaybox}{
  colback=black!3,
  colframe=black!35,
  boxrule=0.4pt,
  sharp corners,
  left=6pt, right=6pt,
  top=4pt, bottom=4pt,
  before skip=6pt, after skip=6pt
}

\usepackage{etoolbox}
\makeatletter
\AtBeginDocument{%
  \patchcmd{\@setref}{\reset@font\bfseries ??}{\reset@font\bfseries\color{red}BROKEN}
    {}{\typeout{** ref patch failed}}%
  \patchcmd{\NAT@citex}{\reset@font\bfseries ?}{\reset@font\bfseries\color{red}BROKEN}
    {}{\typeout{** cite patch 1 failed}}%
  \patchcmd{\NAT@citexnum}{\reset@font\bfseries ?}{\reset@font\bfseries\color{red}BROKEN}
    {}{\typeout{** cite patch 2 failed}}%
}
\makeatother

\usepackage{etoolbox}
\makeatletter
\AtBeginDocument{%
  \patchcmd{\NAT@citex}{\reset@font\bfseries ?}{\reset@font\bfseries\color{red}BROKEN}
    {}{\typeout{** cite patch 1 failed}}%
  \patchcmd{\NAT@citexnum}{\reset@font\bfseries ?}{\reset@font\bfseries\color{red}BROKEN}
    {}{\typeout{** cite patch 2 failed}}%
}
\makeatother

\makeatletter
\AtBeginDocument{%
  \let\NAT@citex@orig\NAT@citex
  \def\NAT@citex[#1][#2]#3{%
    \ifblank{#3}%
      {\PackageWarning{natbib}{EMPTY citation on page \thepage}%
       \NAT@citex@orig[#1][#2]{EMPTY-CITE-KEY}}%
      {\NAT@citex@orig[#1][#2]{#3}}}%
  \let\NAT@citexnum@orig\NAT@citexnum
  \def\NAT@citexnum[#1][#2]#3{%
    \ifblank{#3}%
      {\PackageWarning{natbib}{EMPTY citation on page \thepage}%
       \NAT@citexnum@orig[#1][#2]{EMPTY-CITE-KEY}}%
      {\NAT@citexnum@orig[#1][#2]{#3}}}%
  \ifNAT@numbers
    \let\@citex\NAT@citexnum
  \else
    \let\@citex\NAT@citex
  \fi
}
\makeatother

\title{Beyond Token Savings: A Systematic Study of Context Compression in LLM Agents
}

\author{Ritul Satish, Prasoon Sinha, Akiho Kawada, Neeraja J. Yadwadkar  \\
The University of Texas at Austin\\
}

\iclrfinalcopy
\begin{document}
\raggedbottom
\maketitle

\begin{abstract}

As LLM agents tackle longer tasks, they increasingly compress growing histories of reasoning, actions, and tool outputs. Compression can reduce token use, but it also changes the information available for later decisions. Existing agentic harnesses bundle decisions about what to compress, when to compress, and how much to remove into fixed policies. A systematic characterization is needed to disentangle these decisions and reveal how each affects task success and execution cost. We systematically vary these decisions across three open-weight models on SWE-bench Verified and Terminal-Bench 1.0. Across nearly 35,000 agent runs, we measure task success, token use, end-to-end latency, and estimated cost. We find that fewer tokens need not mean faster or cheaper execution: on Terminal-Bench with Qwen, policies using roughly one-third as many tokens can take 20–80\% longer than the uncompressed agent. Policies with similar overall success can solve different tasks, while the same policy can perform quite differently across models. 
Our results motivate evaluating compression by its effects on agent execution and tailoring policies to the task, model, and workload.
\end{abstract}

\input{sections/01-Introduction}

\input{sections/02-CompDesignSpace}
\input{sections/03-Setup-new}

\input{sections/04-Results-new}

\input{sections/05-Conclusions}

\subsection*{AI use statement}
In this work, we used generative AI tools for to assist with implementing context compression policies, developing experiment and analysis scripts and 
generating plots.
We have not used generative AI tools for to develop the conceptual framework, formulate claims, design the methodology or experiments, and the remaining categories in the ICLR policy are not applicable.

\subsection*{Reproducibility statement}

We document compression implementations and summarization prompts
in Appendix~\ref{app:primitives}, task cohorts and experimental
configurations in Appendix~\ref{app:exp-configs}, and threshold
calibration in Appendix~\ref{app:trigger-threshold-calib}.
Appendix~\ref{app:metrics} specifies metric definitions,
run-inclusion rules, task pairing, and statistical procedures.
Appendix~\ref{app:limits} reports run outcomes and resource-limit
handling to clarify how termination behavior affects the
interpretation of our results.

\bibliography{conference}
\bibliographystyle{conference}

\clearpage
\appendix
\input{sections/06-appendix}

\end{document}

%% file: math_commands.tex
\usepackage{amsmath,amsfonts,bm}

\def\eqref#1{equation~\ref{#1}}

\def\1{\bm{1}}

\DeclareMathAlphabet{\mathsfit}{\encodingdefault}{\sfdefault}{m}{sl}
\SetMathAlphabet{\mathsfit}{bold}{\encodingdefault}{\sfdefault}{bx}{n}



%% file: sections/01-Introduction.tex
\section{Introduction}
\label{sec:intro}

\vspace{-4mm}
\begin{wrapfigure}{R}{2.7in}
  \centering
  \vspace{-1mm}
  \includegraphics[width=\linewidth]{
    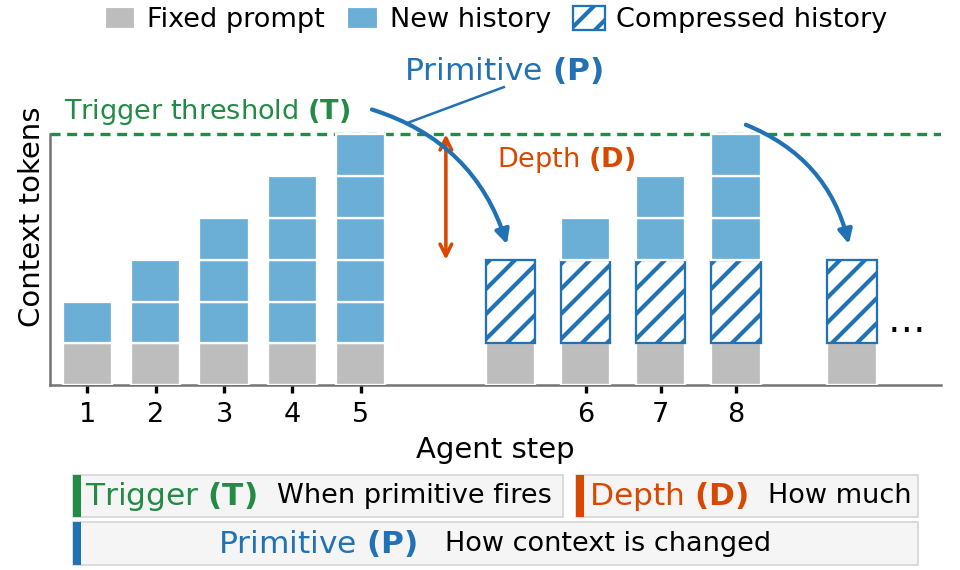
  }
  \vspace{-5mm}
  \caption{Agent context grows across steps and is repeatedly
  compressed by a policy defined by primitive ($P$), trigger ($T$), and depth ($D$).}
  \label{fig:intro-img}
  \vspace{-2mm}
\end{wrapfigure}

Agentic systems increasingly execute long-horizon workflows that interleave LLM reasoning, tool calls, code execution, and environment interaction. As these trajectories grow to hundreds of steps (Figure~\ref{fig:intro-img}), their accumulated history can approach or exceed finite context windows~\citep{hu2025hiagent, packer2023memgpt}. Even before reaching these limits, processing increasingly long contexts raises execution cost and can make relevant information harder for models to use effectively~\citep{tian2025distance, du2025context, liu2024lost, levy2024same}. So, long-running agents need mechanisms for controlling how much of their history is carried forward.

Context compression has emerged as one such mechanism. Modern coding agents selectively remove, rewrite, or summarize portions of an execution trajectory as it grows~\citep{claude_compaction, openai_compaction}. Its impact is already substantial in production: in a trace of 13 million GitHub Copilot sessions, sessions requiring compaction account for 44.2\% of all tokens served, and the median compaction event removes 72.8\% of the existing context~\citep{liu2026agentic}. Compression, however, is not simply a mechanism for saving tokens or fitting within a context window. It changes the information available to subsequent agent decision. Removing information that becomes useful later can alter the agent's trajectory and cause a task to fail that it would otherwise have solved~\citep{liu2026agentic}; compression mechanisms that invoke an LLM can themselves add token and latency overheads.

A context-compression system must decide \emph{how} to compress context, \emph{when} to trigger compression, and \emph{how much} context to remove. Existing systems make different choices along each dimension. Some discard older interactions~\citep{openai_truncation, langchain_trim}, while others summarize them using additional LLM calls~\citep{openai_compaction, claude_compaction}. They also differ in when compression is triggered and how much history is retained~\citep{packer2023memgpt, wang2025openhands}. These choices cannot be considered independently. Compressing early may reduce the context processed by later steps, but it may also cause more compression events. Removing more context can further reduce token usage, but also increases the chance of discarding information needed later.

Because these choices interact, end-to-end evaluations of complete policies are difficult to interpret. A policy may improve task success because of its compression mechanism, its trigger point, the amount of context it retains, or an interaction among these choices. Similarly, two policies with nearly identical aggregate resolve rates may alter the trajectories of entirely different tasks. So, evaluating complete policies leaves a basic question unanswered: \emph{how do the individual decisions that constitute a compression policy shape task success, execution cost, and one another?} Moreover, because models differ in how they use long and compressed contexts, it is unclear whether conclusions obtained for one model transfer to another.

In this work, we systematically characterize context compression as a runtime policy that jointly controls execution cost and the information available to future agent decisions. We vary the compression mechanism, trigger point, and aggressiveness, and organize our study around three questions.

\noindent\textbf{1. What tradeoffs do different compression policies create between task success and execution cost?}
Compression can reduce the context processed during execution, but the resulting savings depend on when compression occurs, how much context is removed, and the cost of compression itself. At the same time, removing useful information can reduce task success. We characterize how these policy choices jointly affect resolve rate, token consumption, and end-to-end latency.

\noindent\textbf{2. How does compression change which tasks an agent solves?} 
Aggregate resolve rate can hide substantial changes in task-level behavior. Two policies may achieve similar overall success rates while solving different tasks: compression can turn some failures into successes while causing previously successful tasks to fail. So, we compare outcomes at the task level to reveal effects that aggregate metrics obscure.

\noindent\textbf{3. Do the effects of compression policies transfer across models?}
A compression policy changes the information presented to the model, so its effect may depend on how that model uses long and compressed contexts. We study whether policies exhibit similar task-success and efficiency tradeoffs across models, or whether effective choices must be tuned to the model executing the task.

To answer these questions, we conduct a controlled, large-scale evaluation of context-compression policies. We examine three policy dimensions, compression mechanism, trigger point, and aggressiveness, while holding the agent loop and task set fixed. We evaluate 65 policies across three open-weight models on SWE-bench Verified and TerminalBench, and repeat each policy--model--task combination three times to quantify run-to-run variability, yielding nearly 35,000 agent executions.

Our study reveals that context compression is not a single efficiency knob: its mechanism, timing, and aggressiveness can change both execution cost and the tasks an agent ultimately solves. We further find that conclusions drawn for one model do not necessarily carry over to another. We quantify these effects in detail in \S\ref{sec:results} and use them to derive implications for the design and evaluation of context-compression policies.

%% file: sections/02-CompDesignSpace.tex
\vspace{-2mm}
\section{Design Space of Compression Policies}
\label{sec:taxonomy}
\vspace{-3mm}

Context compression is already used by widely deployed agents from OpenAI, Anthropic, and Google~\citep{gemini_cli_compression,copilot_compaction,codex_compaction,claude_context_editing}, as well as open-source systems~\citep{openhands_condenser,cline_compaction,pi_compaction}. These systems differ in their implementations, but each must make three fundamental decisions: what information to remove or rewrite, when to compress, and how much context to remove.
Existing systems typically bundle these decisions into particular compression policies, making it difficult to understand which choice is responsible for an observed change in task success or execution cost. We decompose a compression policy along three axes: \emph{primitive}, \emph{trigger}, and \emph{depth}. The primitive determines \emph{how} the trajectory is compressed; the trigger determines \emph{when} compression occurs; and the depth determines \emph{how much} context is removed. Together, one choice along each axis defines a compression policy.
This decomposition lets us vary the three decisions independently and isolate their effects on task success, token consumption, and latency. It also provides a common design space for comparing compression strategies used by existing agents. We describe each axis next.

\vspace{-2mm}
\subsection{Compression Primitives}
\label{subsec:primitive-axis}
\vspace{-2mm}

The primitive determines how information is removed or rewritten during a compression event. Different primitives preserve different parts of the trajectory. For example, truncation removes older agent steps while preserving recent history, tool-result clearing removes older tool outputs while retaining the surrounding interaction, and summarization replaces prior history with a shorter LLM-generated representation. Thus, the choice of primitive determines what evidence, actions, and intermediate state remain available to the agent after compression.

We group primitives into three families according to how they remove information: rule-based primitives delete or clear context with a fixed procedure, LLM-based primitives rewrite context into a shorter representation, and stacked primitives apply both in sequence. Across these families, we study eight primitives drawn from deployed agentic systems(Table~\ref{tab:deployed}). Every primitive operates only on the \emph{compressible window}, which is the accumulated assistant messages and tool results from the trajectory. The system prompt and task instruction are always kept verbatim. Each primitive is defined only by what it removes from the compressible window. The trigger determines when the primitive is applied (\S\ref{subsec:trigger-axis}), while the depth determines how much context it removes (\S\ref{subsec:depth-axis}).

\vspace{-1mm}
\noindent\textbf{Rule-based primitives.}
Rule-based primitives remove information using a fixed procedure and require no additional model invocation. We study two forms used by existing systems.

\vspace{-1mm}
\begin{itemize}[noitemsep,topsep=0pt,leftmargin=*]
\item \emph{Truncation (TR)} removes the oldest messages from the compressible window until the target context size is reached~\citep{openai_truncation,langchain_trim}. It preserves the most recent trajectory verbatim, but discards earlier interactions entirely.
\item \emph{Tool-result clearing (TRC)} replaces older tool outputs with a short placeholder such as \texttt{[output cleared]}, while retaining the surrounding agent interactions and recent tool results~\citep{claude_context_editing}. Unlike truncation, TRC selectively removes a particular source of context growth rather than dropping an entire prefix of the trajectory.
\end{itemize}
\vspace{-1mm}
These primitives illustrate an important distinction. TR decides what to preserve primarily by \emph{recency}, whereas TRC decides by \emph{content type}.

\vspace{-1mm}
\noindent\textbf{LLM-based primitives.}
LLM-based primitives instead preserve prior context by rewriting it into a shorter representation. A summarizer receives some portion of the compressible window and produces a summary that replaces the original text. Unlike rule-based deletion, summarization can retain information from across the compressed history, but it requires an additional model invocation and can omit or alter information during rewriting.
The LLM-based primitives we study vary in two ways: \emph{which portion of the trajectory is summarized} and \emph{how the summary is structured}. A primitive may summarize the entire compressible window or only its older portion, leaving recent interactions verbatim. The generated summary may also be free-form or constrained to preserve predefined categories of information. Unless otherwise stated, we use the same model for agent execution and summarization; we separately evaluate sensitivity to the summarizer model in~\S\ref{app:summarizer}.

\vspace{-1mm}
\noindent\textbf{Stacked primitives.}
Finally, stacked primitives combine rule-based removal with LLM-based summarization. For example, a policy can first clear older tool results and then summarize the remaining history. This mirrors deployed systems that combine multiple forms of context reduction rather than relying on a single operation~\citep{gemini_cli_compression,claude_code_compaction}. Stacking can reduce the amount of text passed to the summarizer, but also changes which information is available when the summary is produced. Thus, these primitive families differ not only in how much context they remove, but in \emph{which information survives compression and in what form}: verbatim recent history, selected content types, or a rewritten representation of prior interactions.

\vspace{-2mm}
\subsection{Compression Triggers}
\label{subsec:trigger-axis}
\vspace{-3mm}

The trigger determines when compression occurs, shaping how much
history accumulates and how often the agent intervenes to reduce
it. We study two strategies: \emph{threshold-based} triggers tied
to context size and \emph{step-based} triggers tied to agent steps.

\vspace{-1mm}
\noindent\textbf{Threshold trigger ($Th$).}
Compression fires when the trajectory exceeds a token threshold and can recur as context grows afterward. Lower thresholds trigger earlier and potentially more often; higher thresholds allow more history to accumulate. The threshold value is a trigger parameter, not a separate policy axis. We evaluate tight,
primary, and loose settings for Qwen and Devstral, and only the primary setting for GLM (\S\ref{sec:setup}), to study sensitivity
to when compression occurs.

\vspace{-1mm}
\noindent\textbf{Step trigger ($St$).}
A step trigger instead invokes compression as the agent executes, independent of the current context size. In our evaluation, compression runs after every agent step. Thus, the number of compression events is determined by trajectory length rather than by how quickly the trajectory accumulates tokens. We evaluate the step trigger with tool-result clearing, which allows old tool outputs to be removed continuously as they accumulate.

These two triggers impose different notions of \emph{when context has become worth compressing}: threshold triggers respond to context size, whereas step triggers respond to trajectory progress.

\vspace{-2mm}
\subsection{Compression Depth}
\label{subsec:depth-axis}
\vspace{-2mm}

Once compression is triggered, the depth determines how much of the current context is removed. This creates another tradeoff. Shallow compression preserves more of the existing trajectory, but leaves less room for subsequent context growth and may cause compression to fire again soon. Deeper compression creates more headroom, but removes more information at each event.
We define compression depth, $D$, as 
$D = 1 - \frac{\texttt{target tokens}}{\texttt{current tokens}}$,
where \texttt{current tokens} is the size of the compressible window when compression begins and \texttt{target tokens} is its target size after compression. Thus, $D=0.5$ reduces the compressible window to half its pre-compression size, while larger values indicate more aggressive compression. We evaluate $D \in \{0.3,0.5,0.7\}$, with $D=0.5$ as the canonical setting.
We distinguish \emph{depth-tunable} primitives, such as truncation
and summarization, whose retained history or summary size can be
targeted, from \emph{depth-invariant} primitives, such as
tool-result clearing, whose effective depth depends on the
operation and available content. We sweep depth only for
depth-tunable primitives.

\vspace{-2mm}
\subsection{Compression Policies}
\label{subsec:policy-notation}
\vspace{-2mm}

A compression policy combines a primitive, a trigger, and, when
applicable, a depth. We denote a policy as $P/T/D$, where $P$
identifies the primitive, $T$ the trigger, and $D$ the compression
depth. For example, $SU\text{-}p/Th/0.5$ denotes partial
summarization triggered by a token threshold and targeting a
50\% reduction in the compressible window.
Appendix~\ref{app:deployed} maps representative deployed systems
onto this design space.

%% file: sections/03-Setup-new.tex
\vspace{-2mm}
\section{Characterization Methodology}
\label{sec:setup}
\vspace{-3mm}

We use the design space from \S\ref{sec:taxonomy} to systematically characterize how compression primitives, triggers, and depths affect agent task success and execution cost. Our methodology is designed to separate the effects of these policy choices while controlling the agent harness, task, and model within each comparison. We evaluate policies across two agentic coding benchmarks and three open-weight models, and repeat each configuration to account for run-to-run variability.

\vspace{-1mm}
\noindent\textbf{Benchmarks and Tasks.} 
We evaluate compression policies on two coding benchmarks, SWE-bench Verified~\citep{jimenez2024swebench} and Terminal-Bench 1.0~\citep{merrill2026terminal}. In SWE-bench Verified, the agent resolves a real GitHub issue by modifying a repository, and held-out tests evaluate the resulting patch. In Terminal-Bench, the agent completes command-line tasks through multi-step interaction with a terminal. For SWE-bench, we follow prior work~\citep{wang2025openhands,hou2024large} and use 100 issues from Django, SymPy, and scikit-learn. For Terminal-Bench, we use 40 tasks. The threshold and depth sweeps use subsets of 25 and 15 of these tasks, respectively. Appendix~\ref{app:tasks} lists the cohorts and their composition.

\vspace{-1mm}
\noindent\textbf{Agent harness.}
We use mini-swe-agent~\citep{yang2024swe}, whose minimal agent loop adds no memory or context-management mechanism that could confound compression, and we apply benchmark-specific step and time limits (Appendix~\ref{app:harness}).

\vspace{-1mm}
\noindent\textbf{Models.} 
We evaluate Qwen3.5-35B-A3B, Devstral-Small-2-24B, and
GLM-4.7-Flash to test whether compression effects transfer
across models. Models are served locally using
vLLM~\citep{kwon2023efficient},
with the agent model also performing summarization by default.
Appendix~\ref{app:harness} provides model and serving details.

\vspace{-1mm}
\noindent\textbf{Trigger-Threshold Calibration.}
Threshold-based policies require choosing the context size at which compression is triggered. Setting the threshold only at the model's maximum context window would evaluate compression primarily as a response to imminent context overflow. Our goal is broader: to characterize when compression becomes useful during normal agent execution, including trajectories that never approach the model's context limit.

So, we calibrate trigger thresholds from the context demand of uncompressed trajectories. For each model and benchmark, we first execute the tasks without compression and record the peak per-call context size reached by each trajectory. We then define three trigger thresholds settings from this distribution: \emph{tight}, \emph{primary}, and \emph{loose}, corresponding to its 5th, 15th, and 25th percentiles, respectively. Lower thresholds expose more trajectories to compression and tend to trigger it earlier, while higher thresholds restrict compression to trajectories with larger context demands.

This calibration serves two purposes. First, it places the trigger relative to the context demands of each model and workload rather than using the same arbitrary token count across models with different context behavior. Second, the three threshold settings let us study whether conclusions about a compression policy depend on when it is invoked. We determine all thresholds from uncompressed runs before evaluating the corresponding compressed configurations. Appendix~\ref{app:trigger-threshold-calib} reports the resulting thresholds and context-demand distributions.

\vspace{-1mm}
\noindent\textbf{Characterization Matrix. }
We organize our experiments into two complementary tracks (Table~\ref{tab:exp-grid}). The first provides a broad comparison across compression policies, while the second varies individual policy parameters to isolate their effects.

\begin{table}[t]
\centering\scriptsize
\setlength{\tabcolsep}{2.5pt}
\begin{tabular}{@{}l@{\hspace{4pt}}l ccccc cccccc cc@{}}
\toprule
 & & \multicolumn{5}{c}{Depth-tunable} & \multicolumn{6}{c}{Depth-invariant} & \multicolumn{2}{c}{Reference} \\
\cmidrule(lr){3-7}\cmidrule(lr){8-13}\cmidrule(lr){14-15}
\multicolumn{2}{@{}l}{Trigger}   & TR & SU & SU-p & SS & SS-p & TRC & TRC+SU & TRC+SS & OTRC+TR & OTRC+SU-p & OTRC+SS-p & FC & OTRC \\
\multicolumn{2}{@{}l}{Threshold $T$} & $Th$ & $Th$ & $Th$ & $Th$ & $Th$ & $Th$ & $Th$ & $Th$ & $St{+}Th$ & $St{+}Th$ & $St{+}Th$ & -- & $St$ \\
\midrule
\multirow{3}{*}{Threshold} & Tight   & $\circ\circ\circ$ & $\circ\circ\circ$ & $\circ\circ\circ$ & $\circ\circ\circ$ & $\circ\circ\circ$ & $\circ$ & $\circ$ & $\circ$ & $\circ$ & $\circ$ & $\circ$ & \multirow{3}{*}{$\bullet$} & \multirow{3}{*}{$\bullet$} \\
 & Primary & $\circ\bullet\circ$ & $\circ\bullet\circ$ & $\circ\bullet\circ$ & $\circ\bullet\circ$ & $\circ\bullet\circ$ & $\bullet$ & $\bullet$ & $\bullet$ & $\bullet$ & $\bullet$ & $\bullet$ & & \\
 & Loose   & $\circ\circ\circ$ & $\circ\circ\circ$ & $\circ\circ\circ$ & $\circ\circ\circ$ & $\circ\circ\circ$ & $\circ$ & $\circ$ & $\circ$ & $\circ$ & $\circ$ & $\circ$ & & \\
\bottomrule
\end{tabular}

\vspace{1mm}
\centering
\textit{Legend:} $\bullet$ Full comparison, \quad
$\circ$ Controlled sweeps (task sets defined in \S\ref{sec:setup}).

\vspace{-2mm}
\caption{\small{Policy grid, run for each model at its three threshold settings from \S\ref{sec:setup}. In each cell the marks read left to right as $D = 0.3, 0.5, 0.7$. Depth-invariant primitives do not depend on $D$ and therefore have a single mark per row. FC and OTRC has no threshold, so each is one cell. Every cell is run three times per task.}}
\label{tab:exp-grid}
\end{table}

\vspace{-1mm}
\emph{Full comparison.}
We evaluate all compression primitives on the full benchmark task sets using each model's primary threshold and the canonical compression depth $D=0.5$. This comparison captures how substantially different compression strategies affect task success, token consumption, and latency under a common operating point. We also include the full-context reference and the step-triggered tool-result-clearing policy.

\vspace{-1mm}
\emph{Controlled sweeps.}
We use the smaller stratified task subsets to vary trigger threshold and compression depth. For depth-tunable primitives, we evaluate $D \in \{0.3,0.5,0.7\}$ at each of the tight, primary, and loose thresholds. Primitives whose effective depth is determined by the operation itself are evaluated once per threshold. These sweeps let us separate the effect of \emph{when} compression occurs from \emph{how aggressively} it removes context without requiring the full combinatorial grid on every task.

Every task--policy--model configuration is repeated three times to capture run-to-run variability. Table~\ref{tab:exp-grid} summarizes the resulting evaluation matrix; Appendix~\ref{app:grid} (Table~\ref{tab:grid}) reports the complete configuration and run counts for both benchmarks.

\vspace{-1mm}
\noindent\textbf{Metrics and analysis.}
We measure task success, token consumption, billed input cost,
and end-to-end latency over three runs per task--policy--model
configuration, examining both aggregate and task-level outcomes.
Appendix~\ref{app:metrics} details metric definitions, inclusion
rules, and statistical comparisons; Appendix~\ref{app:limits}
reports run outcomes and resource limits.

%% file: sections/04-Results-new.tex
\vspace{-2mm}
\section{Results}
\label{sec:results}
\vspace{-3mm}

\subsection*{Question \#1 (Q1). What tradeoffs do different compression policies create between task success and execution cost?}
\label{subsec:results-q1}
\vspace{-2mm}

\begin{figure}[t]
  \centering
  \includegraphics[width=.9\textwidth]{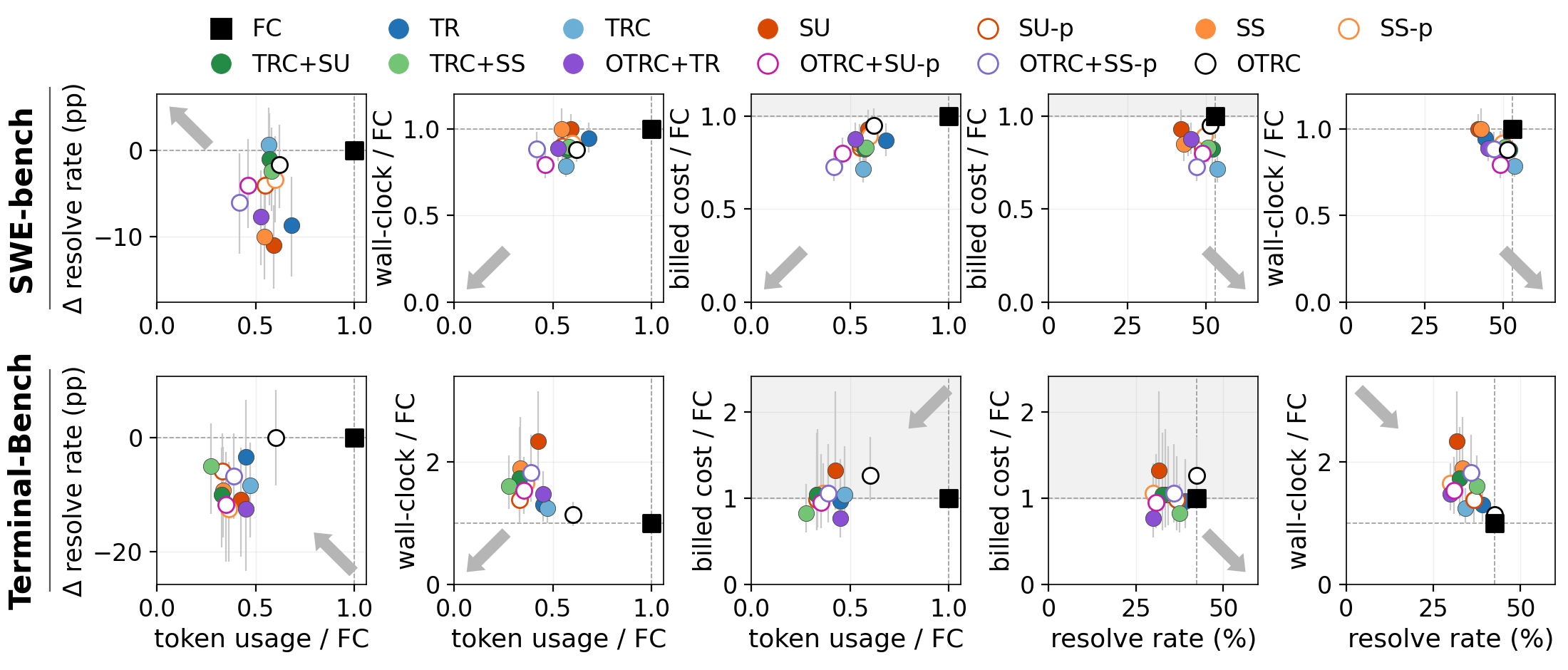}
  \vspace{-4mm}
  \caption{\small{Qwen's success and resource tradeoffs on SWE-bench (top) and Terminal-Bench (bottom). Resource metrics are paired ratios to full context on completed runs; success includes all runs. Gray shading marks higher billed cost, and dotted diagonals mark equal proportional token and bill savings. Vertical bars show 95\% task-bootstrap intervals.
  }}
  \label{fig:q1-qwen-overview}
\end{figure}

Figure~\ref{fig:q1-qwen-overview} shows Qwen3.5-35B-A3B on
SWE-bench (top) and Terminal-Bench (bottom). Resource metrics are ratios of task-averaged values to Full Context (FC) on shared eligible tasks, denoted by ``/ FC''; $0.5$ means half the FC value. The x-axis shows relative token usage; the columns show resolve-rate change in percentage points over all runs,
relative latency, and relative billed cost (Appendix~\ref{app:bill}), respectively.

\noindent\textbf{Observation \#1: Context compression consistently reduces token usage, but can trade off success, latency, and billed cost in different ways.}
\vspace{-2mm}

\noindent\textit{Tradeoffs observed on SWE-bench.}
Figure~\ref{fig:q1-qwen-overview} shows that every compression policy reduces token usage relative to full context. On SWE-bench, policies use roughly $0.4$--$0.7\times$ the tokens, while success ranges from $11.0$ percentage points (pp) below full context to $0.7$ points above it. TRC mostly matches full-context success while reducing token usage, latency, and billed cost to $0.57\times$, $0.79\times$, and $0.71\times$, respectively. SU uses a similar $0.59\times$ the tokens but reduces success by $11.0$ pp.

\vspace{-1mm}
\noindent\textit{Similar token usage, different outcomes.}
These differences persist across policies with comparable token consumption. On SWE-bench, policies using roughly $0.56$--$0.63\times$ the tokens of full context span about 12 percentage points in resolve rate, $0.79$--$1.0\times$ in latency, and $0.71$--$0.95\times$ in billed cost. Even similar success rates can mask different execution costs: TRC and OTRC achieve resolve rates of 53.7\% and 51.3\%, respectively, yet their billed costs are $0.71\times$ and $0.95\times$ full context, and their latencies are $0.79\times$ and $0.88\times$.

\vspace{-1mm}
\noindent\textit{Trade-offs observed on Terminal-Bench.}
On Terminal-Bench, most policies reduce token usage more aggressively, to roughly $0.25$--$0.4\times$ full context, but 11 of the 12 reduce success by 5--13 percentage points. All policies also take longer than full context, with latency ranging from roughly $1.05\times$ to $1.8\times$ for most policies. Billed cost nevertheless decreases for 9 of the 12 policies, with several reaching roughly $0.4$--$0.7\times$ full context. We suspect that the workloads contribute to this difference: SWE-bench agents can recover working context from code and patches, whereas Terminal-Bench's varied workflows may depend more on intermediate observations and action history.

Token savings alone therefore do not identify an effective or efficient policy. The same reduction in tokens can accompany different success rates and execution costs, motivating a closer look at what determines billed cost and latency.

\noindent\textbf{Observation \#2: Token savings do not yield proportional billed-cost or latency savings.
}
\vspace{-2mm}

\noindent\textit{Observed mismatch.}
The middle and right columns of Figure~\ref{fig:q1-qwen-overview} show how large this mismatch can be. On SWE-bench, using roughly $0.6\times$ the tokens of full context still incurs $0.71$--$0.95\times$ the billed cost and $0.79$--$1.0\times$ the latency. On Terminal-Bench, policies using roughly one-third of the tokens can take $1.2$--$1.8\times$ as long, and some even cost more than full context. Compression can therefore save tokens without delivering comparable monetary savings, and can increase runtime despite much smaller prompts. Total token usage combines the tokens processed per call with the number of calls, but does not distinguish the factors that determine billed cost and latency.

\vspace{-1mm}
\noindent\textit{Billed cost: fresh vs cached input.}
Billed cost depends on how many input tokens remain eligible for the cached rate. Under the assumed prefix-cached pricing, cached input tokens cost one-tenth as much as fresh input tokens (Appendix~\ref{app:bill}). A policy that frequently rewrites the trajectory can remove tokens while invalidating much of the reusable prefix. It may therefore process fewer tokens but charge more of them at the fresh-input rate, producing a similar or even higher estimated bill.

\vspace{-1mm}
\noindent\textit{Latency: call duration and execution length.}
Total latency depends on both the duration and number of calls, as well as tool execution and compression overhead. Shorter prompts can make individual model calls faster, yet additional agent calls, tool execution, and compression overhead can offset those savings. Step-triggered policies make the largest reductions in tokens per step but require 10--27\% more calls across models. On Qwen, faster calls more than offset the additional calls; on Devstral, calls themselves become slower, compounding the increase in execution length. In contrast, stacked threshold policies reduce token usage by 22--55\% while keeping call counts near full context and latency at $0.88$--$1.01\times$.

These results explain why token savings alone cannot predict execution cost: compression also changes which tokens receive the cached rate, how many calls the agent makes, and how much time each step takes.

\noindent\textbf{Observation \#3: Trigger threshold and depth shape different execution tradeoffs.}
\vspace{-2mm}

\begin{figure*}[t]
    \centering
    \includegraphics[width=\textwidth]{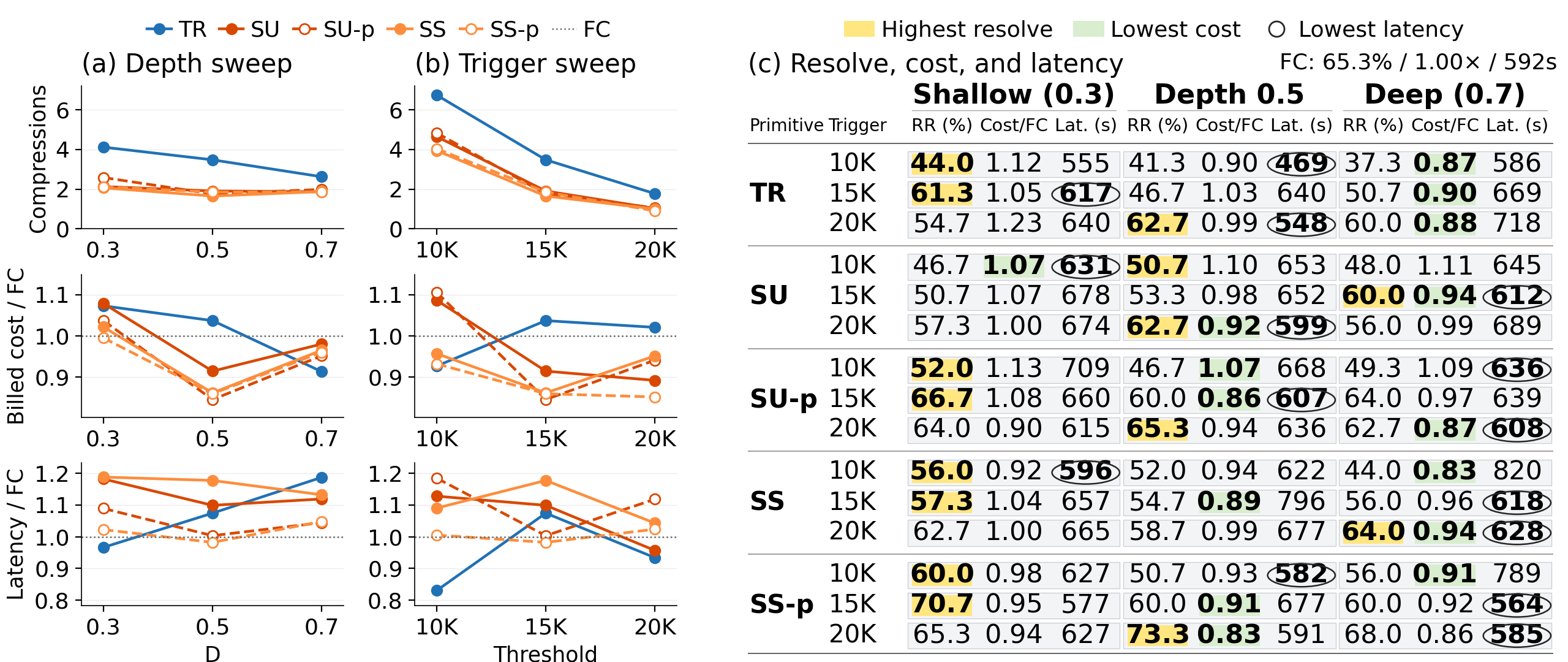}
    \vspace{-7mm}
    \caption{\small{Trigger and depth tradeoffs on SWE-bench (Qwen).
    Left: compression frequency, input cost, and latency.
    Right: resolve rate, cost relative to FC, and latency per setting;
    yellow, green, and circles mark best resolve, cost, and latency
    across depths. Left resource curves exclude time-capped attempts;
    the table includes all attempts.}}
    \label{fig:q1-tuning}
    \vspace{-5mm}
\end{figure*}

\noindent\textit{Experiment.}
To separate when compression occurs from how much history it removes, we vary the trigger threshold at fixed depth and compare depths at fixed thresholds. Figure~\ref{fig:q1-tuning} reports the Qwen SWE-bench sweeps on 25 tasks, with three attempts per task: thresholds of 10K, 15K, and 20K tokens and depths $D\in\{0.3,0.5,0.7\}$. We measure compression frequency, resolve rate, billed input cost, and end-to-end latency. 

\vspace{-1mm}
\noindent\textit{Threshold effects.}
Delaying compression generally helps in these sweeps. Increasing the threshold from 10K to 20K at $D=0.5$ reduces average compression frequency from $4.85$ to $1.15$ events per attempt and improves resolve rate across all tested primitives. For SU, resolve rate rises from $50.7\%$ to $62.7\%$, while input cost drops from $1.10\times$ to $0.92\times$ full context and latency falls from 653 to 599 seconds. Retaining more history before compression can therefore reduce repeated interventions while improving success and efficiency.

\vspace{-1mm}
\noindent\textit{Depth effects.}
The effect of depth depends on the primitive. At a 15K threshold, increasing TR's depth from $0.3$ to $0.7$ lowers compression frequency and input cost, but resolve rate drops from $61.3\%$ to $50.7\%$ and latency increases. SU shows the opposite trend: resolve rate rises from $50.7\%$ to $60.0\%$ while cost and latency decrease. Excluding time-capped attempts preserves these directions: TR loses $15.4$ percentage points of success, whereas SU gains $7.4$ points. Across the 15 primitive--threshold combinations, shallow compression gives the highest resolve rate in eight cases; deep compression gives the lowest cost in eight and the lowest latency in seven.

\vspace{-1mm}
\noindent\textit{Differences across benchmark suites.}
Although both benchmarks use the same calibration procedure, larger tool outputs on Terminal-Bench rebuild context more quickly, triggering compression more frequently than on SWE-bench.
On shared solved tasks, Qwen with TR compresses about 9 times per run on Terminal-Bench, compared with fewer than 2 on SWE-bench. For five policies, total token usage remains at or above $0.94\times$ full context despite shorter prompts, while summarization calls account for $28$--$46\%$ of solved-run wall-clock time for the 4 summarization primitives.

\vspace{-2mm}
\begin{takeawaybox}
\textbf{Takeaway 1.}
No single compression policy dominates across task success
and execution cost. So, the choice of a policy needs to reflect the workload
and the user's priorities for success, latency, and cost.
\end{takeawaybox}

\vspace{-1mm}
\subsection*{Question \#2 (Q2). How does compression change which task an agent can solve?}
\label{subsec:results-q2}
\vspace{-2mm}

Q1 shows that some policies approximately match full-context
success at lower execution cost, but may solve different tasks.
We therefore compare tasks gained and lost relative to FC,
counting a task as solved when at least two of three attempts
succeed. Appendix~\ref{app:task-specificity} shows the breakdown
for Qwen's 100 tasks, separating the 47 that FC misses from the
53 it solves; Appendix~\ref{app:task-statistics} details outcome
handling.

\begin{figure}[t]
    \centering
    \includegraphics[width=\linewidth]{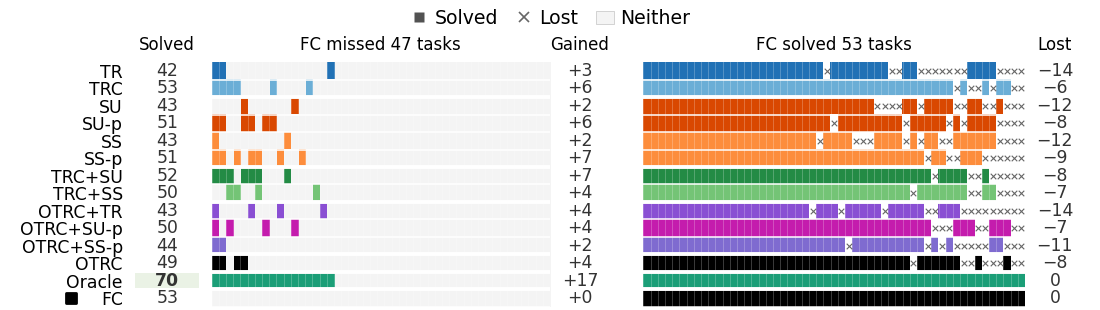}
    \vspace{-7mm}
    \caption{\small{Qwen on SWE-bench. Task coverage relative to FC, with gains, losses, and total solved.}}
    \label{fig:q2-task-coverage}
\end{figure}

\noindent\textbf{Observation \#4: Similar overall performance can mask substantial task-level differences.}
\vspace{-1mm}

Figure~\ref{fig:q2-task-coverage} shows substantial task turnover under compression.
Across compression policies, 2--7 of the 47 tasks that full context misses become newly solved, while 6--14 of the 53 tasks solved by full context become unsolved.
These gains and losses can largely offset each other.
For example, TRC+SU solves 52 tasks compared with 53 for full context, but newly solves 7 tasks while losing 8 previously solved tasks.
SS-p shows a similar pattern, gaining 7 tasks and losing 9, for a total of 51 solved tasks.
In contrast, TR gains only 3 tasks while losing 14, reducing the total to 42.
Thus, approximately matching full-context success does not mean preserving its task coverage: gains on other tasks can mask failures on tasks that full context previously solved.

\vspace{-2mm}
\begin{takeawaybox}
\textbf{Takeaway 2.}
The best compression policy varies by task, motivating task-aware customization rather than a single fixed policy.
\end{takeawaybox}

\subsection*{Question \#3 (Q3). Do compression policies effects transfer across models?}
\label{subsec:results-q3}
\vspace{-2mm}

\begin{figure}[t]
    \centering
    \includegraphics[width=\linewidth]{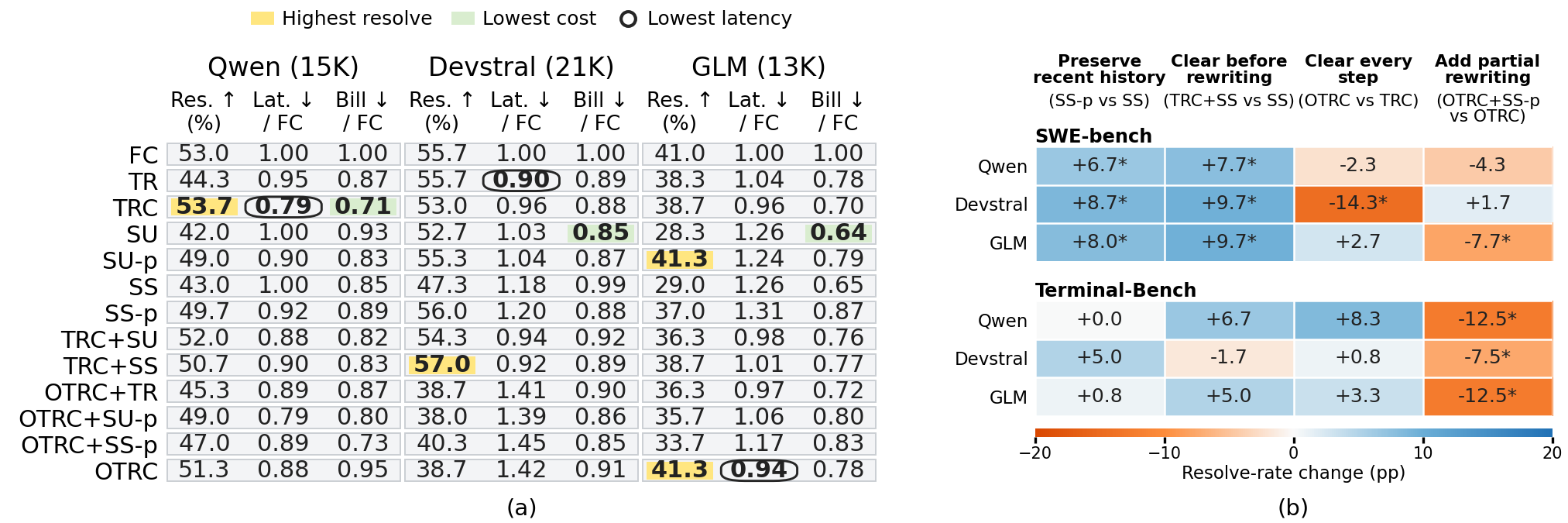}
    \vspace{-8mm}
    \caption{\small{
    (a) SWE-bench policy comparisons: darker cells indicate better
    within-model ranks; outlines mark best values. Latency and billed
    input cost use uncapped, task-paired ratios to FC.
    (b) Resolve-rate changes (first policy minus second);
    $^*$ marks 95\% task-bootstrap intervals excluding zero.
    Primary thresholds; three attempts per task
    (SWE-bench: 100 tasks; Terminal-Bench: 40).
    }}
    \label{fig:q3}
    \vspace{-3mm}
\end{figure}

Q1 and Q2 show that compression changes both execution cost and
which tasks an agent solves. We now examine whether these effects
transfer across models. Figure~\ref{fig:q3} compares policy rankings
for Qwen, Devstral, and GLM, then tests whether individual design
choices have consistent effects despite differences in the
best-performing policies.

\noindent\textbf{Observation \#5: Compression-policy effects can change substantially across models.}

\vspace{-1mm}
\noindent\textit{Same policy, different effects.}
Figure~\ref{fig:q3}a shows that the same compression policy can have very different effects on resolve rate, latency, and billed cost depending on the model. 
For example, OTRC performs well with Qwen and GLM, reaching 51.3\% and 41.3\% resolve rate while reducing latency to $0.88\times$ and $0.94\times$ full context, respectively. 
With Devstral, however, the same policy drops resolve rate to 38.7\% and increases latency to $1.42\times$ full context. Thus, the impact of a given compression policy on both success and efficiency can change substantially across models.

\vspace{-1mm}
\noindent\textit{Model-dependent policy rankings.}
The policy that performs best also changes across models. With Qwen, TRC achieves the highest resolve rate at $53.7\%$, the lowest billed cost at $0.71\times$, and the lowest latency at $0.79\times$ full context. 
With Devstral, TRC+SS achieves the highest resolve rate at 57.0\%, while TR has the lowest latency at $0.90\times$ and SU the lowest billed cost at $0.85\times$. 
With GLM, OTRC and SU-p achieve the highest resolve rate at 41.3\%, while SU has the lowest billed cost at $0.64\times$ but increases latency to $1.26\times$ and reduces resolve rate to 28.3\%. 
The relative ordering of policies can even reverse across models: OTRC improves resolve rate over TR by $7.0$ percentage points with Qwen, but reduces it by $17.0$ points with Devstral. 
Therefore, no single compression policy performs best across models, and a policy that is effective for one model can be substantially worse for another.

\vspace{-1mm}
\noindent\textit{Configuration sensitivity.} Configuration effects also differ between Qwen and Devstral in the controlled sweeps, which vary threshold and depth while holding the primitive fixed.
In our smaller ablation cohorts, increasing SS-p's SWE-bench trigger threshold from 10K to 15K to 20K increases Qwen's resolve rate from 50.7\% to 60.0\% to 73.3\%, while Devstral remains roughly unchanged at 72--73\% across its three thresholds. 
Compression depth shows a similar pattern: decreasing SU-p's depth from $D=0.7$ to $0.5$ to $0.3$ on Terminal-Bench improves Qwen from 33.3\% to 40.0\% to 46.7\%, but changes Devstral from 37.8\% to 22.2\% to 37.8\%, respectively.

These comparisons establish model-dependent responses to compression, but do not isolate their causes. In particular, they do not distinguish differences in information retention, recovery from removed context, or execution behavior; thresholds are also calibrated separately for each model. Explaining the ranking reversals requires further controlled analysis.

\vspace{-1mm}
\noindent\textbf{Observation \#6: While policy behavior is model-dependent, some design choices have consistent effects across models. 
}
\vspace{-1mm}

Although policy rankings differ across models, some design choices have consistent effects. Figure~\ref{fig:q3}b compares these choices through paired policy comparisons; results below are reported in the order Qwen, Devstral, and GLM.

\vspace{-1mm}
\noindent\textit{Shared design benefits.}
On SWE-bench, preserving recent history through partial structured rewriting (SS-p vs. SS) improves resolve rate by 6.7, 8.7, and 8.0 percentage points, respectively. Clearing tool outputs before structured rewriting (TRC+SS vs. SS) improves resolve rate by 7.7, 9.7, and 9.7 points. Both changes benefit all three models, providing consistent design guidance on this workload despite their different overall policy rankings.

\vspace{-1mm}
\noindent\textit{Model-dependent effects.}
Other design choices do not show this consistency. Clearing tool outputs every step rather than only at the threshold (OTRC vs. TRC) changes SWE-bench resolve rate by $-2.3$, $-14.3$, and $+2.7$ percentage points. The same change therefore substantially hurts Devstral while having smaller, oppositely signed effects on Qwen and GLM.

\vspace{-1mm}
\noindent\textit{Workload limits.}
Consistency across models does not guarantee transfer across workloads. Adding partial structured rewriting to per-step clearing (OTRC+SS-p vs. OTRC) reduces Terminal-Bench resolve rate by 12.5, 7.5, and 12.5 percentage points, while its effects on SWE-bench are mixed. 

\vspace{-2mm}
\begin{takeawaybox}
\textbf{Takeaway 3.}
Compression policies are not one-size-fits-all across models, but their effects are not entirely model-specific either. Some design choices remain consistently beneficial across models, suggesting opportunities to combine reusable policy components with model- and workload-aware adaptation where needed.
\end{takeawaybox}

%% file: sections/05-Conclusions.tex
\vspace{-2mm}
\section{Conclusion}
\vspace{-3mm}
We characterized the effects of context compression across two coding benchmarks and three open-weight models. Saving tokens did not reliably reduce latency or estimated input cost; the same compression policy could help on some tasks and hurt on others; and compression policy effects varied across models, although some design choices improved success for all three on SWE-bench. These findings motivate judging context compression by the executions it enables and tailoring policies to the task, model, and workload, with the goal of retaining the information an agent needs to finish efficiently.

%% file: sections/06-Appendix.tex
\raggedbottom

\newcommand{\ablationpart}[3]{%
    \centering
    \begin{subfigure}{\linewidth}
        \centering
        \includegraphics[width=\linewidth]{figures/appendix/depth_trigger_ablation/#1}
        \caption{#2}
        \label{#3}
    \end{subfigure}}

\newcommand{\ablationgroup}[6]{%
\begin{figure}[H]
    \ablationpart{#1_tight.png}{Tight threshold (#3)}{fig:app-#6-tight}
\end{figure}
\begin{figure}[H]
    \ContinuedFloat
    \ablationpart{#1_primary.png}{Primary threshold (#4)}{fig:app-#6-primary}
\end{figure}
\begin{figure}[H]
    \ContinuedFloat
    \ablationpart{#1_loose.png}{Loose threshold (#5)}{fig:app-#6-loose}
    \caption{Depth and threshold ablation for #2.}
    \label{fig:app-#6}
\end{figure}}

\newcommand{\taskmapgroup}[5]{%
\begin{figure}[H]
    \centering
    \begin{subfigure}{\linewidth}
        \centering
        \includegraphics[width=\linewidth]{figures/appendix/task_specificity/#1_#3_D0_3.png}
        \caption{$D=0.3$}
        \label{fig:app-taskmap-#5-#3-d03}
    \end{subfigure}
    \vspace{2mm}
    \begin{subfigure}{\linewidth}
        \centering
        \includegraphics[width=\linewidth]{figures/appendix/task_specificity/#1_#3_D0_5.png}
        \caption{$D=0.5$}
        \label{fig:app-taskmap-#5-#3-d05}
    \end{subfigure}
    \vspace{2mm}
    \begin{subfigure}{\linewidth}
        \centering
        \includegraphics[width=\linewidth]{figures/appendix/task_specificity/#1_#3_D0_7.png}
        \caption{$D=0.7$}
        \label{fig:app-taskmap-#5-#3-d07}
    \end{subfigure}
    \caption{Task coverage relative to full context for #2 at the #3 threshold (#4).}
    \label{fig:app-taskmap-#5-#3}
\end{figure}}

\newcommand{\taskmapfig}[7]{%
\begin{figure}[H]
    \centering
    \includegraphics[width=\linewidth,height=0.92\textheight,keepaspectratio]{figures/appendix/task_specificity/#1_#3.png}
    \caption{\textbf{#4 threshold (#5).} Task coverage relative to full context for #2 (threshold/depth sweeps, #7 tasks).}
    \label{fig:app-taskmap-#6-#3}
\end{figure}}

\section{Appendix}
\label{sec:appendix}

Appendix~\ref{app:related-works} reviews related work.
Appendix~\ref{app:primitives} details the compression implementations,
summarizer prompts, and mappings to deployed systems.
Appendices~\ref{app:exp-configs} and~\ref{app:trigger-threshold-calib}
describe the experimental configurations and threshold calibration.
Appendix~\ref{app:metrics} defines the metrics and statistical procedures.
Appendix~\ref{app:extended-results} presents threshold and depth sweeps,
task-level gains and losses, summarizer sensitivity, run outcomes
and resource limits, and changes in execution length.

\subsection{Related Work}
\label{app:related-works}
\noindent\textbf{Prior agentic serving characterization.}
Long agent trajectories can become difficult to serve effectively even before reaching the model's context limit. 
Prior work shows that models make weaker use of distant information and often incur increasing token overhead as context grows~\citep{packer2023memgpt, hu2025hiagent, liu2024lost, tian2025distance, du2025context}. 
Recent characterization studies examine related effects at different levels: multi-turn performance degradation in LLMs~\citep{laban2026llms}, end-to-end execution~\citep{kim2026dynamicreasoning, raj2026cpucentric, chang2026agentsysbench}, resource usage~\citep{yang2026agora, zheng2026agentcgroup}, and serving behavior of agents~\citep{kondrashov2026aries, kim2026gaiatrace, yuan2026kairos, yuan2026workloadchar}, and production-scale workload, context growth, and cache behavior in GitHub Copilot~\citep{liu2026agentic}. 
Together, these works establish that growing agent context affects both model behavior and serving efficiency. 
However, they characterize workloads under a given context-management mechanism rather than systematically studying the compression policy itself. 
Our work instead varies \emph{what} is compressed, \emph{when} compression is triggered, and \emph{how aggressively} history is compressed, and measures how these choices change task success, execution length, cost, and latency.

\noindent\textbf{Context management inside the agent loop.}
Modern agents increasingly manage growing histories by truncating observations, summarizing earlier interactions, or combining both~\citep{codex_compaction, claude_context_editing, gemini_cli_compression, copilot_compaction, openhands_condenser, cline_compaction}. 
These mechanisms are already consequential in production: across 13 million Copilot sessions, sessions that invoke compaction account for 44\% of served tokens, and the median compaction removes 73\% of the context~\citep{liu2026agentic}.

Recent work proposes compression techniques that substantially reduce retained context while often preserving task success~\citep{kang2025acon, chen2026coact, lindenbauer2025complexity, xiao2026reducing}. 
However, compression can also alter the execution itself: agents may take additional steps, revisit discarded information, or incur extra model calls to construct summaries~\citep{min2026toward, lindenbauer2025complexity}. Observational studies of deployed coding agents also identify
failures to carry forward earlier constraints and
decisions~\citep{tang2026coding}. Although these failures are
not attributed specifically to compression, they motivate
examining what information context-management policies preserve.
Consequently, evaluating compression only by the number of tokens removed can miss important downstream effects. 
We build on this observation by treating compression as a policy with separable design choices (primitive, trigger, and depth) and studying how each shapes the resulting agent trajectory across models and task environments.

\noindent\textbf{Serving efficiency and end-to-end cost.}
Prompt size is also an incomplete proxy for serving cost. Modern serving systems reuse KV state ~\citep{kwon2023efficient,qin2025mooncake} and shared prompt prefixes~\citep{zheng2024sglang,gim2024promptcache,srivatsa2025preble} to reduce the cost of repeated context, while commercial APIs similarly discount cached input tokens~\citep{claude_prompt_caching,openai_prompt_caching,gemini_context_caching}. 
Compression can disrupt this reuse: rewriting earlier history may invalidate a cached prefix~\citep{liu2025lmcache}, while summarization itself introduces additional inference. 
Production Copilot traces observe reduced cache reuse around compaction~\citep{liu2026agentic}, and TokenPilot proposes prefix-stable compaction specifically to preserve reuse~\citep{xu2026tokenpilot}. 
As a result, reducing prompt tokens does not necessarily reduce either billed cost or end-to-end latency.

Our study connects these effects in a controlled evaluation of the compression policy itself. 
Rather than treating compression efficiency as the size of the compressed prompt, we evaluate the full execution it induces, including task outcome, token usage, number of agent steps, serving cost, and latency.

\subsection{Primitive Implementation}
\label{app:primitives}
\paragraph{Shared rules.}
The system prompt and initial task instruction are retained verbatim; compression operates on the remaining history. Before each agent call, threshold policies count message-content tokens using \texttt{tiktoken} (\texttt{cl100k\_base}) and fire when the estimate exceeds $B$. This estimate excludes model-specific chat-template overhead.

\paragraph{Truncation (TR).}
TR removes the oldest whole messages until the target is reached, always retaining the final compressible message. It does not preserve action--response pairs, and an oversized final message can leave the context above target.

\paragraph{Tool-result clearing (TRC).}
TRC replaces older tool responses with short placeholders, retaining surrounding assistant messages and the three most recent eligible responses. It skips existing clearing and summary markers. Clearing proceeds oldest first until the target is reached; if insufficient, standalone TRC falls back to TR. This fallback can remove recent responses that were protected from clearing.

\paragraph{Summarization (SU and SS).}
Both replace the compressible history with one user-role summary. SU requests free-form prose covering the task objective, progress, files examined or modified, observations, errors, decisions, and remaining work. SS instead requires five ordered sections: Task, Files Modified, Files Examined, Execution Anchors, and Current State. Both prompts emphasize preserving exact paths and relevant errors; SS additionally requests exact commands, test results, and symbol names and prohibits unsupported additions. The summarizer receives a role-tagged transcript of the compressible history, including any previous summary, but not the protected prefix.

\paragraph{Partial summarization (SU-p and SS-p).}
Partial variants allocate half the available token allowance to a verbatim recent tail and half to a summary of the older head. The tail is the longest contiguous suffix of whole messages that fits its allowance; it may be empty. The head uses the corresponding SU or SS prompt. If the reconstructed context exceeds the target, TR removes the oldest messages, potentially including the summary.

\paragraph{Step-triggered clearing (OTRC).}
Before each call, OTRC clears the response preceding the four most recent tool responses in an alternating action--response history. The implementation replaces the ninth message from the end once at least five calls and ten messages are present. This positional rule also applies after stacked rewrites, without checking message roles or summary markers. Standalone OTRC has no token threshold.

Summary prompts request approximately $0.75$ words per allocated token, with a minimum allowance of 50 tokens. By default, the summarizer uses the agent model, temperature $0.2$, and a 4096-token output cap. Summary length is a soft target: SU has no length fallback, and SS's TR fallback cannot shorten its sole retained summary message. The exact SU and SS prompt templates are reproduced below.

\subsubsection{Summarizer prompt templates}
\label{app:summary-prompts}
The system and user messages below reproduce the templates in \texttt{memory.py}, with line wrapping for readability. At runtime, \texttt{\{target\_words\}} is replaced by the requested word count and \texttt{\{history\_text\}} by the role-tagged transcript. SU-p and SS-p use the same templates with only the older head as the transcript.

\noindent\textbf{SU: System message.}
\begin{quote}
\small
\begin{verbatim}
You are summarizing an agent's work history to free up context
window space. Your output will replace the conversation history
— the agent will only see your summary, so it must be complete
enough to continue the task.
\end{verbatim}
\end{quote}

\noindent\textbf{SU: User message.}
\begin{quote}
\small
\begin{verbatim}
Summarize the following agent conversation history in
approximately
{target_words} words. Your summary MUST include all of:
1. Current task objective and progress made so far
2. Files examined and any modifications made (include exact file
paths)
3. Key observations, errors encountered, and decisions taken
4. Current state — what has been done and what remains

Preserve exact file paths, error messages, and code snippets
that are likely still relevant. Be factual and concise.

{history_text}
\end{verbatim}
\end{quote}

\noindent\textbf{SS: System message.}
\begin{quote}
\small
\begin{verbatim}
You are compressing an agent's working memory to free up context
window space. Your output will REPLACE the entire conversation
history — the agent will only see your summary. It must contain
everything needed to continue the task without any other
context.
\end{verbatim}
\end{quote}

\noindent\textbf{SS: User message.}
\begin{quote}
\small
\begin{verbatim}
Produce a structured summary of the agent conversation below in
approximately
{target_words} words total. Use EXACTLY this format and section
order:

[CONTEXT SUMMARY]
## Task
<One sentence: what the task is asking for.>

## Files Modified
<Each file the agent has written or edited. Format: path — what
changed and why. If none yet, write 'None'.>

## Files Examined
<Each file the agent has read. Format: path — key observation.
If none yet, write 'None'.>

## Execution Anchors
<Exact commands run, test results, error messages, or shell
output that the agent will need to reference going forward.
Quote verbatim where possible.>

## Current State
<What has been done, what has NOT been done, and the immediate
next step.>
[END CONTEXT SUMMARY]

Rules:
- Preserve exact file paths, line numbers, error strings, and
symbol names.
- Do not speculate or add information not present in the
history.
- Do not add any text outside the [CONTEXT SUMMARY] block.

Conversation history:
{history_text}
\end{verbatim}
\end{quote}

\subsubsection{Compression in deployed systems}
\label{app:deployed}

Table~\ref{tab:deployed} maps representative deployed compression
mechanisms onto our primitive--trigger--depth design space.
These mappings identify the closest policy analogues;
implementation details may differ from our evaluated policies.

\begin{table}[t]
\centering
\scriptsize
\setlength{\tabcolsep}{3pt}
\renewcommand{\arraystretch}{1.15}
\begin{tabularx}{\textwidth}{
  >{\raggedright\arraybackslash}p{1.45cm}
  >{\raggedright\arraybackslash\hsize=1.15\hsize}X
  >{\raggedright\arraybackslash\hsize=0.95\hsize}X
  >{\raggedright\arraybackslash\hsize=0.9\hsize}X
  c
  c}
\toprule
System
  & What is removed ($P$)
  & When it fires ($T$)
  & How much ($D$)
  & User sets
  & $P/T/D$ \\
\midrule
OpenHands\newline\citeyearpar{openhands_condenser}
  & Older half of the log $\rightarrow$ free-form summary
  & $Th$ at 80 events (default) or a token cap
  & $D = 0.5$; keeps first few and latest events
  & $P$, $T$, $D$
  & $SU\text{-}p/Th/0.5$ \\
\addlinespace
Gemini CLI\newline\citeyearpar{gemini_cli_compression}
  & Tool results truncated; older 70\% $\rightarrow$ structured summary
  & $Th$ at 0.5 of the context window
  & $D = 0.7$ (fixed); last 30\% survives
  & $T$
  & $TRC{+}SS\text{-}p/Th/0.7$ \\
\addlinespace
Claude Code\newline\citeyearpar{claude_code_compaction}
  & Tool outputs cleared; history $\rightarrow$ structured summary
  & $Th$ at the context limit (0.97 on a 1M window)
  & $D \rightarrow 1$; nothing kept verbatim
  & $T$
  & $TRC{+}SS/Th/1.0$ \\
\bottomrule
\end{tabularx}
\vspace{-2mm}
\caption{Three deployed compression mechanisms placed in the three-axis design space. In the $P$
column $\rightarrow$ reads as replaced by, and a semicolon separates the stages of a stacked
primitive. $Th$ denotes a threshold trigger, given as an event count or a window fraction. $D$ is the
fraction of context removed, after both stages. The last column gives the closest analogue in our
notation rather than an exact match. All three fire on a threshold, and their depths run from $0.5$
to $1$, so our canonical depth sits on a deployed default. See \S\ref{sec:taxonomy}.    }
\label{tab:deployed}
\end{table}

\subsection{Experimental Configurations}
\label{app:exp-configs}

\paragraph{Tasks and evaluation subsets.}
\label{app:tasks}
SWE-bench Verified evaluates repository-level bug fixing on real GitHub issues. Our 100 issues come from Django (34), SymPy (34), and scikit-learn (32), which cover web development, symbolic mathematics, and machine learning. For Terminal-Bench 1.0 we use the \texttt{terminal-bench-core} 0.1.1 dataset. Its 40 tasks cover build and environment setup, data processing, machine-learning workflows, security, system administration, and version control. The main comparisons use all selected tasks. The threshold and depth sweeps use subsets of 25 SWE-bench and 15 Terminal-Bench tasks to limit the cost of evaluating many configurations (Table~\ref{tab:task-cohorts}).

\begin{table}[h]
\centering\small
\begin{tabular}{lrr}
\toprule
Benchmark & Main comparison & Threshold/depth sweeps \\
\midrule
SWE-bench Verified & 100 & 25 \\
Terminal-Bench & 40 & 15 \\
\bottomrule
\end{tabular}
\caption{Number of tasks in each evaluation. Sweep tasks are subsets of the main comparison tasks. The SWE-bench sweep subset contains 9 Django, 8 SymPy, and 8 scikit-learn issues.}
\label{tab:task-cohorts}
\end{table}

\paragraph{Harness and serving.}
\label{app:harness}
We use a modified mini-swe-agent harness with three open-weight
models: Qwen3.5-35B-A3B, Devstral-Small-2-24B, and
GLM-4.7-Flash. The models span a dense transformer (Devstral),
a mixture-of-experts transformer (GLM), and a mixture-of-experts
model combining linear and full attention (Qwen), allowing us
to examine whether compression effects transfer across models.

All models are served locally through
vLLM~\citep{kwon2023efficient}, on four A100 80GB GPUs for SWE-bench and four RTX A6000 48GB GPUs for Terminal-Bench, using temperature 0.2 and a 4096-token output cap (Table~\ref{tab:env}).

\paragraph{Experiment grid.}
\label{app:grid}
Table~\ref{tab:grid} summarizes the full comparison for all three models and the controlled sweeps for Qwen and Devstral on both benchmarks. GLM is evaluated only at the primary threshold and fixed depth $D=0.5$, alongside the FC and OTRC references. Model-specific trigger thresholds are given in Appendix~\ref{app:trigger-threshold-calib}.

\begin{table}[t]
\centering
\begingroup
\scriptsize
\setlength{\tabcolsep}{3pt}
\renewcommand{\arraystretch}{1.15}
\resizebox{\linewidth}{!}{%
\begin{tabular}{l ccccc cccccc cc}
\toprule
& \multicolumn{5}{c}{Depth-tunable} & \multicolumn{6}{c}{Depth-invariant} & \multicolumn{2}{c}{Reference} \\
\cmidrule(lr){2-6}\cmidrule(lr){7-12}\cmidrule(lr){13-14}
Trigger threshold & TR & SU & SU-p & SS & SS-p & TRC & TRC+SU & TRC+SS & OTRC+TR & OTRC+SU-p & OTRC+SS-p & FC & OTRC \\
Trigger $T$ & $Th$ & $Th$ & $Th$ & $Th$ & $Th$ & $Th$ & $Th$ & $Th$ & $St{+}Th$ & $St{+}Th$ & $St{+}Th$ & -- & $St$ \\
\midrule
Tight & $\circ\circ\circ$ & $\circ\circ\circ$ & $\circ\circ\circ$ & $\circ\circ\circ$ & $\circ\circ\circ$ & $\circ$ & $\circ$ & $\circ$ & $\circ$ & $\circ$ & $\circ$ & -- & -- \\
Primary & $\circ\bullet\circ$ & $\circ\bullet\circ$ & $\circ\bullet\circ$ & $\circ\bullet\circ$ & $\circ\bullet\circ$ & $\bullet$ & $\bullet$ & $\bullet$ & $\bullet$ & $\bullet$ & $\bullet$ & -- & -- \\
Loose & $\circ\circ\circ$ & $\circ\circ\circ$ & $\circ\circ\circ$ & $\circ\circ\circ$ & $\circ\circ\circ$ & $\circ$ & $\circ$ & $\circ$ & $\circ$ & $\circ$ & $\circ$ & -- & -- \\
No threshold & -- & -- & -- & -- & -- & -- & -- & -- & -- & -- & -- & $\bullet$ & $\bullet$ \\
\bottomrule
\end{tabular}%
}
\par\smallskip
\textit{Legend:} $\bullet$ Full comparison; $\circ$ Controlled sweeps; -- not applicable.\\
SWE-bench: 100 / 25 tasks; Terminal-Bench: 40 / 15 tasks (full / sweeps).\\
$\bullet$: all three models; $\circ$: Qwen and Devstral only.
\endgroup
\caption{Experiment grid for both benchmarks. GLM includes only the filled marks; Qwen and Devstral include both filled and open marks. For depth-tunable policies, the three marks read left to right as $D=0.3,0.5,0.7$; the full comparison uses the primary threshold and $D=0.5$. Depth-invariant policies use one fixed configuration per threshold, as described in Appendix~\ref{app:primitives}. FC and standalone OTRC are evaluated once per task repetition without an operative token threshold. Each marked configuration targets three runs per task. Full-comparison runs on the sweep subsets also provide the corresponding primary-threshold comparisons.}
\label{tab:grid}
\end{table}

The summarizer comparison uses the 25-task SWE-bench subset and Qwen's 15K threshold to compare self-summarization with Qwen3.5-9B and Gemma-4-12B for SU and TRC+SU (Appendix~\ref{app:summarizer}).

\begin{table}[t]
\centering
\caption{Software environment and inference settings. Unless noted, values are shared across Qwen3.5-35B-A3B, Devstral-24B, and GLM-4.7-Flash.}
\label{tab:env}
\begin{tabular}{ll}
\toprule
\textbf{Setting} & \textbf{Value} \\
\midrule
\multicolumn{2}{l}{\textit{Software}} \\
Python              & 3.10.12 \\
CUDA                & 12.3 (Qwen, Devstral) / 12.9 (GLM) \\
PyTorch             & 2.10.0 (Qwen, Devstral) / 2.13.0 (GLM) \\
Transformers        & 4.57.6 (Qwen, Devstral) / 5.16.1 (GLM) \\
vLLM                & 0.17.1 (Qwen, Devstral) / 0.28.0 (GLM) \\
mini-swe-agent      & 2.2.6 \\
swe-bench           & 4.1.0 \\
terminal-bench-core & 0.1.1 (Terminal-Bench 1.0) \\
Harbor              & 0.20.0 \\
\midrule
\multicolumn{2}{l}{\textit{Inference}} \\
GPUs                & 4$\times$ A100 80GB (SWE-bench) / 4$\times$ RTX A6000 48GB (Terminal-Bench) \\
dtype               & bf16 \\
Tensor parallel     & 4 \\
max-model-len       & 102,400 (Qwen) / 65,536 (Devstral, GLM) \\
max-num-seqs        & 64 \\
Concurrency         & 16 (SWE-bench) / 4 (Terminal-Bench) \\
Temperature         & 0.2 \\
\bottomrule
\end{tabular}
\end{table}

\subsection{Trigger-Threshold Calibration}
\label{app:trigger-threshold-calib}

We calibrate thresholds separately for each benchmark and model using uncompressed trajectories. For each trajectory, we record its peak context: the largest server-reported prompt-token count across its calls. We select tight, primary, and loose thresholds near the 5th, 15th, and 25th percentiles of the peak-context distribution, rounding to the nearest 1K tokens. This places compression relative to the context demands of each workload: lower thresholds trigger earlier and affect more tasks. The full comparison uses the primary threshold; controlled sweeps use all three thresholds for Qwen and Devstral. GLM uses only the primary threshold.

Table~\ref{tab:trigger-thresholds} lists the thresholds. Qwen's SWE-bench settings predate the percentile rule and correspond approximately to the 5th, 16th, and 28th percentiles. Calibration uses server-reported counts, while runtime checks use the token estimate described in Appendix~\ref{app:primitives}; exact crossing rates can therefore differ.

\begin{table}[t]
\centering\small
\begin{tabular}{llrrr}
\toprule
Benchmark & Model & Tight & Primary & Loose \\
\midrule
SWE-bench Verified & Qwen3.5-35B-A3B      & 10K & 15K & 20K \\
                   & Devstral-Small-2-24B & 17K & 21K & 24K \\
                   & GLM-4.7-Flash        & 10K  & 13K & 15K  \\
\midrule
Terminal-Bench     & Qwen3.5-35B-A3B      & 2K & 3K & 4K \\
                   & Devstral-Small-2-24B & 3K & 4K & 7K \\
                   & GLM-4.7-Flash        & -- & 3K & -- \\
\bottomrule
\end{tabular}
\caption{Trigger thresholds by benchmark and model, in tokens (K = 1,000). On Terminal-Bench, GLM is evaluated only at the primary threshold.}
\label{tab:trigger-thresholds}
\end{table}

\subsection{Metrics and statistical comparisons}
\label{app:pairing}
\label{app:task-statistics}
\label{app:metrics}
Each task--policy--model configuration is repeated three times.
Task success is determined by the benchmark's native evaluator.
Token consumption includes prompt and output tokens from both
agent and compression calls; latency measures wall-clock time
from trajectory start to termination. Billed input cost is
estimated as described in Appendix~\ref{app:bill}.
We examine aggregate performance and individual-task outcomes,
and compare policy effects relative to each model's full-context
baseline when assessing transfer across models.

\paragraph{Success and task-level comparisons.}
Resolve rate averages each task's success fraction across three runs. Runs terminated by resource limits without a successful outcome count as unresolved. For task gains and losses, we consider a task solved when at least two of its three runs succeed. A gain is a task solved by the policy but not FC; a loss is the reverse. These comparisons use the full cohorts (100 SWE-bench tasks and 40 Terminal-Bench tasks), require three recorded attempts per task and policy, and treat missing verdicts as unresolved. Step comparisons use only successful attempts on tasks solved by both policies, averaging steps within each task before comparing task means.

\paragraph{Resource comparisons.}
Main comparisons of token consumption, cost, and latency exclude runs that reach the time limit (1500 seconds on SWE-bench or the task-specific Harbor timeout on Terminal-Bench). Runs that reach only the step limit are included. Resource accounting requires at least two recorded prompt-token counts per run. We average eligible runs within each task and retain tasks with measurements for both the policy and FC. The reported ratio divides the policy's average across these tasks by FC's average. The controlled-sweep table in Figure~\ref{fig:q1-tuning} (right) includes all attempts in its cost and latency averages; the curves on the left exclude time-capped runs.

\paragraph{Uncertainty.}
We report 95\% confidence intervals by repeatedly sampling tasks with replacement. Each sampled task retains all its runs for both the policy and FC. We recalculate the success difference or resource and step ratios for each sample.

\subsubsection{Token consumption}
\label{app:tokens}
Token consumption is the total number of prompt and output tokens recorded over a run, including the summarization calls to the model for compression. Peak context measures the largest prompt in a single call; cumulative consumption sums usage across calls. A trigger threshold therefore does not bound total token consumption.

\begin{figure}[h]
\centering
\includegraphics[width=0.8\linewidth]{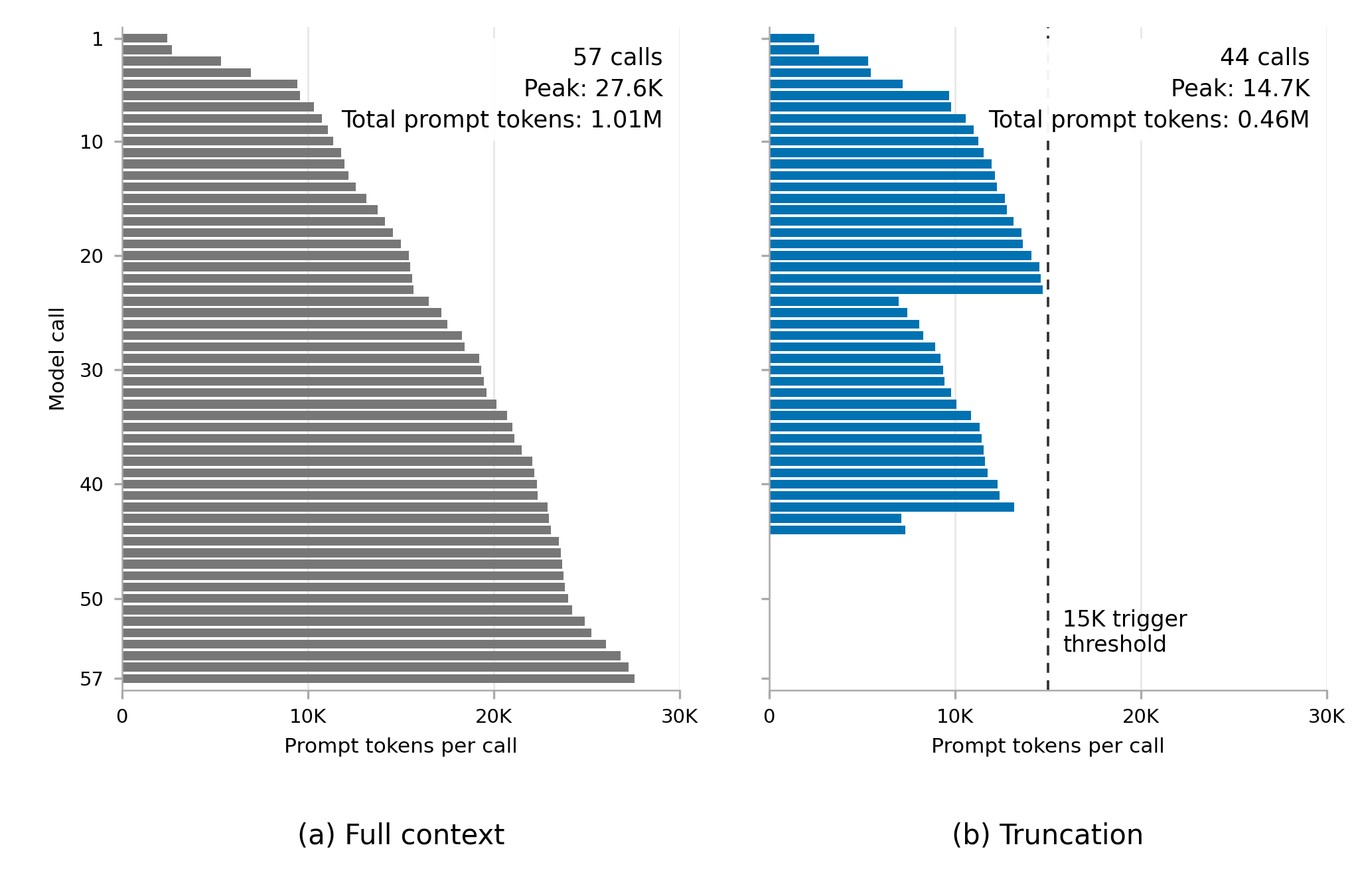}
\caption{Prompt tokens per model call for Qwen3.5-35B-A3B on scikit-learn 13142, run 1: full context (left) and truncation with a 15K trigger threshold (right). Each bar represents one call; cumulative prompt consumption is the sum of the bar lengths.}
\label{fig:budget-explainer}
\end{figure}

Figure~\ref{fig:budget-explainer} illustrates this distinction for two runs of the same task: shorter prompts and fewer calls both reduce cumulative prompt consumption.

\subsubsection{Billed cost calculation}
\label{app:bill}
We estimate billed input cost in units of one fresh input token:
\begin{equation}
\widehat{C}_{\mathrm{input}}
= N_{\mathrm{fresh}} + 0.1\,N_{\mathrm{cached}} + N_{\mathrm{summary}}
\label{eq:input-cost}
\end{equation}
where $N_{\mathrm{fresh}}$ and $N_{\mathrm{cached}}$ are the reconstructed fresh and cached agent input-token counts, and $N_{\mathrm{summary}}$ is the total summarizer input-token count. Cached input is charged at one-tenth the fresh-input rate; summarizer input is treated as fresh. Output tokens contribute to token consumption but are excluded from this input-cost metric.

We estimate fresh and cached input-token counts from successive prompt sizes and recorded OTRC events, using 16-token blocks. For prompt reductions of at least $\max(1000,0.25P)$ tokens, where $P$ is the preceding prompt size, tokens beyond the estimated stable prefix are counted as fresh. For OTRC events, tokens from the cleared message onward are counted as fresh; ordinary prompt growth is also charged as fresh input. For smaller unexplained reductions, we use the conservative reconstruction that charges the suffix beyond the stable prefix as fresh. This is an estimate under the assumed cache discount, not a measured provider bill or cache-hit rate; Qwen was served with prefix caching disabled.

\subsection{Extended Results}
\label{app:extended-results}
We report trigger-threshold and depth sweeps, task-level gains and losses, and sensitivity to the summarizer model.

\subsubsection{Trigger-threshold and depth sweeps}
\label{app:results}
Figures~\ref{fig:app-qwen-swe}--\ref{fig:app-devstral-swe} extend the main comparison to the controlled-sweep subsets. Each subfigure fixes one trigger threshold and compares success, token consumption, latency, and billed input cost across compression depths.

\ablationgroup{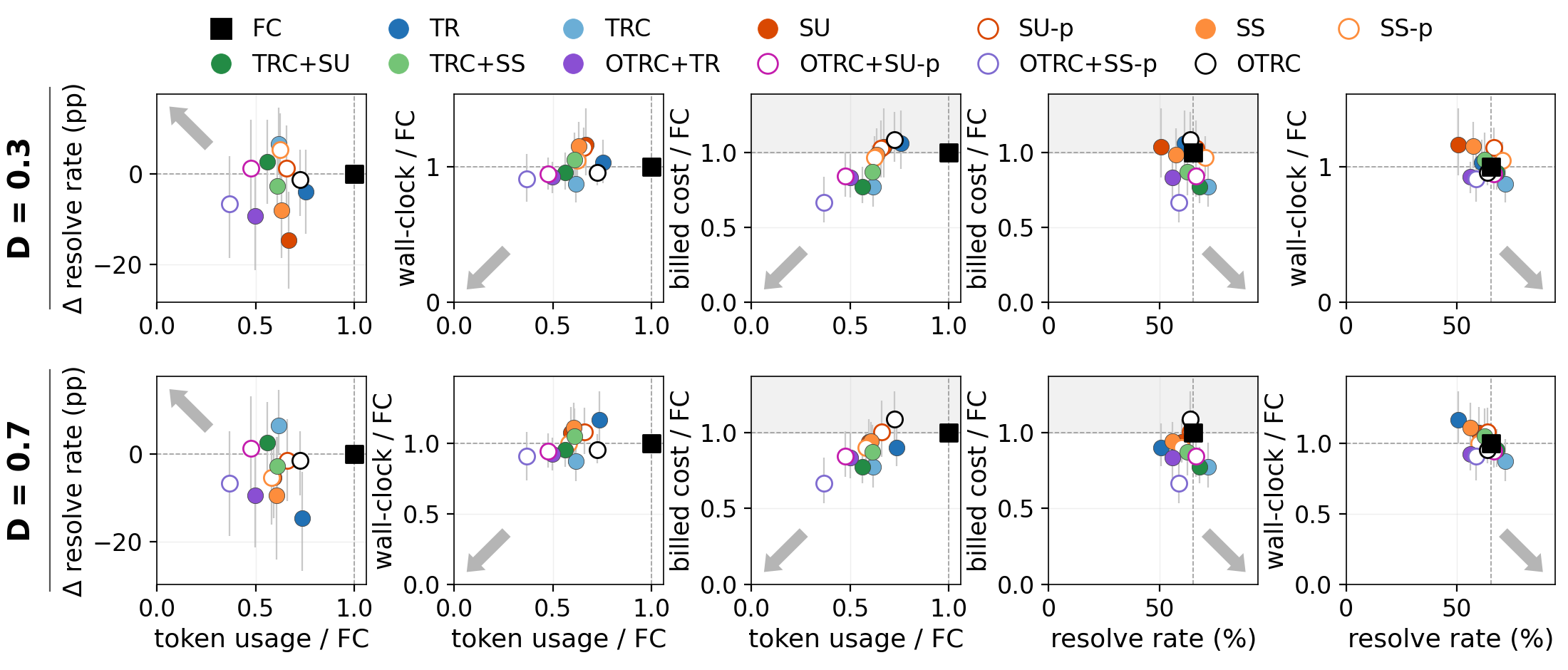}{Qwen3.5-35B-A3B on SWE-bench}{10K}{15K}{20K}{qwen-swe}
\ablationgroup{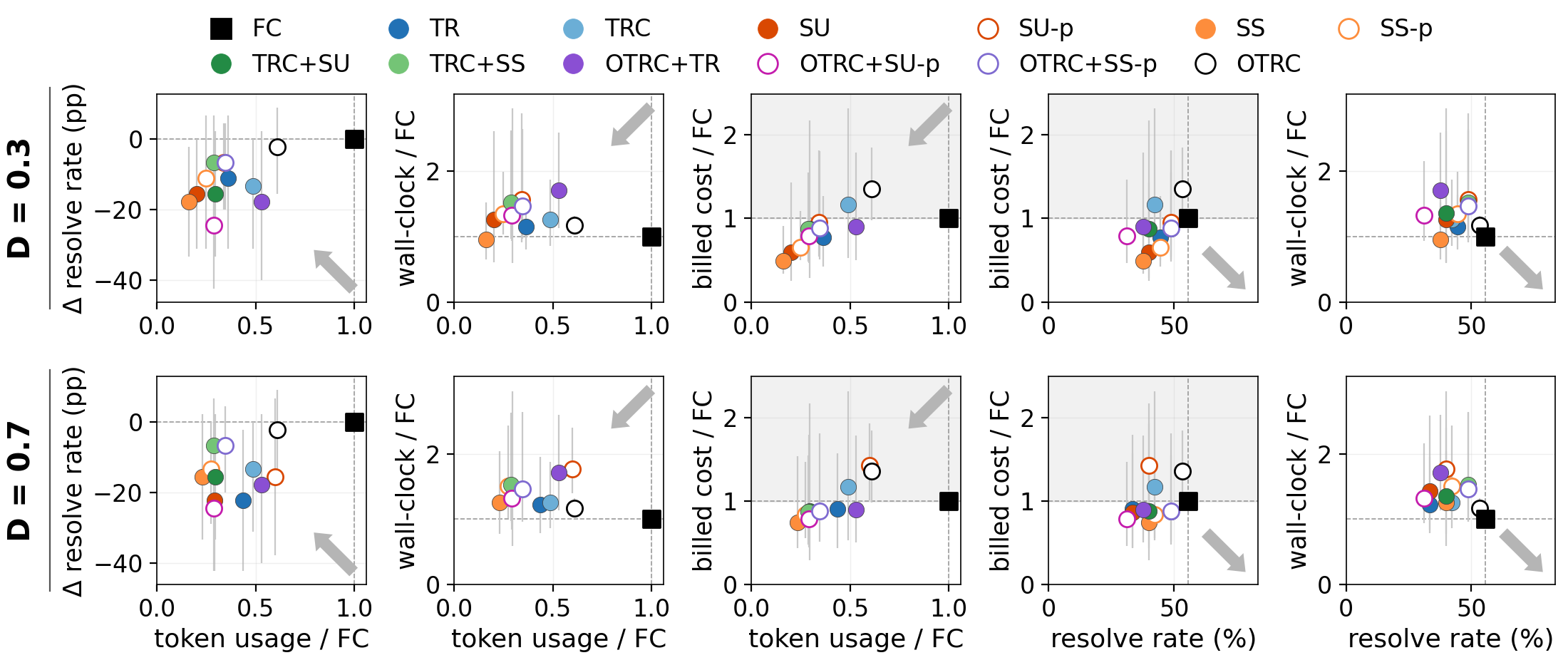}{Qwen3.5-35B-A3B on Terminal-Bench}{2K}{3K}{4K}{qwen-tb}
\ablationgroup{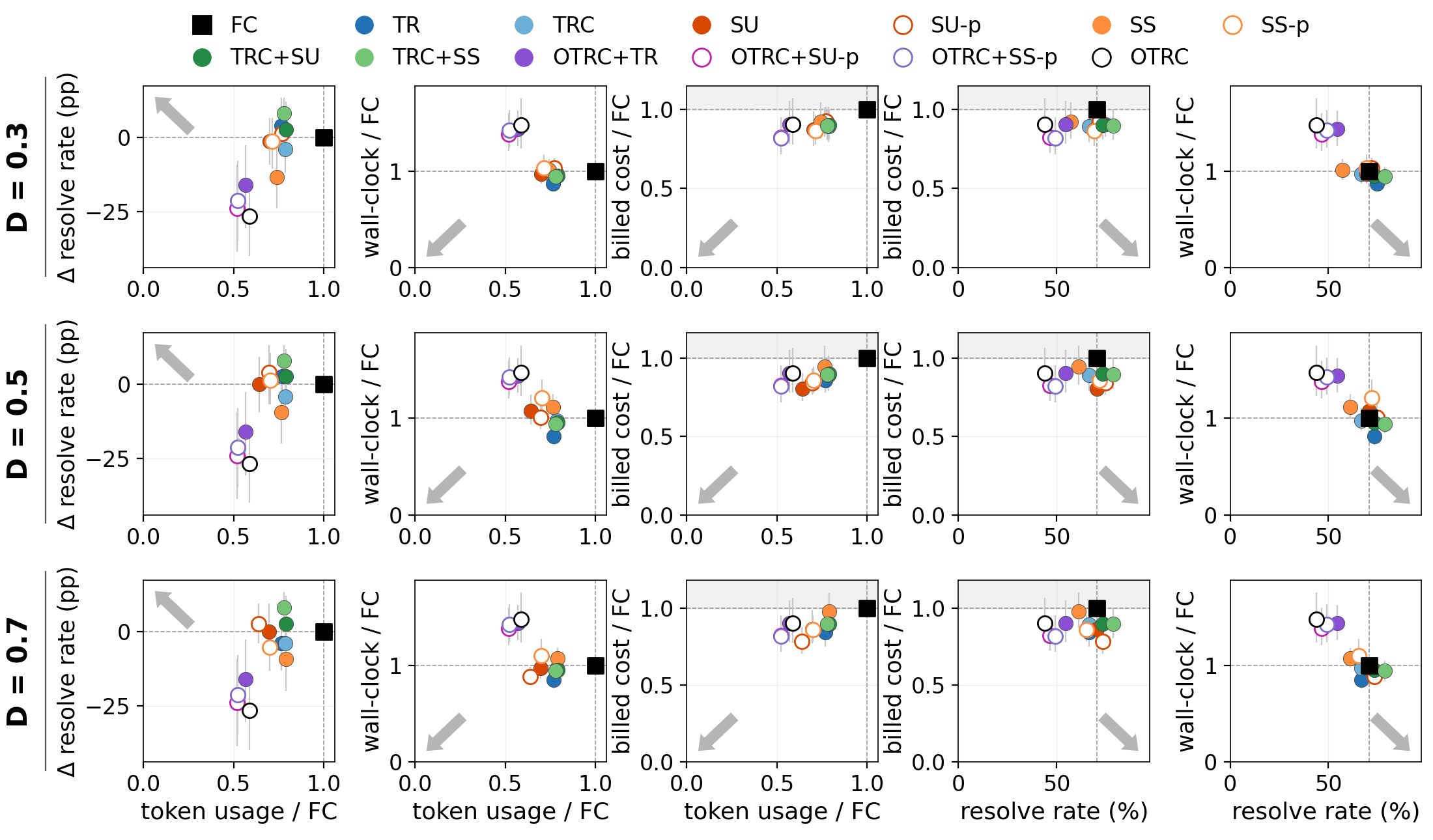}{Devstral-Small-2-24B on SWE-bench}{17K}{21K}{24K}{devstral-swe}

\subsubsection{Task-level gains and losses}
\label{app:task-specificity}
Figures~\ref{fig:app-taskmap-qwen-swe-tight}--\ref{fig:app-taskmap-qwen-swe-loose} extend Figure~\ref{fig:q2-task-coverage} to the 25-task SWE-bench subset across trigger thresholds and compression depths. Each panel shows which tasks Qwen gains or loses relative to FC, using the success criterion in Appendix~\ref{app:task-statistics}.
\taskmapfig{qwen35b_swe_task_map}{Qwen3.5-35B-A3B on SWE-bench}{tight}{Tight}{10K}{qwen-swe}{25}
\taskmapfig{qwen35b_swe_task_map}{Qwen3.5-35B-A3B on SWE-bench}{primary}{Primary}{15K}{qwen-swe}{25}
\taskmapfig{qwen35b_swe_task_map}{Qwen3.5-35B-A3B on SWE-bench}{loose}{Loose}{20K}{qwen-swe}{25}

\subsubsection{Summarizer model sensitivity}
\label{app:summarizer}
\begin{figure}[t]
\centering
\includegraphics[width=\linewidth]{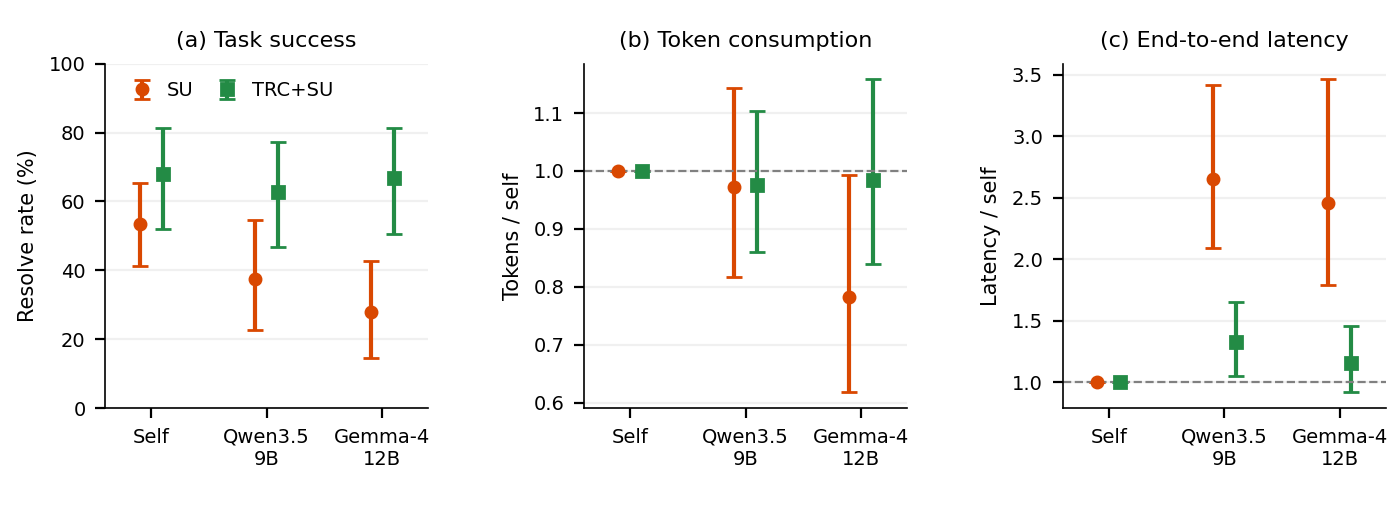}
\caption{Summarizer sensitivity with a fixed Qwen agent on SWE-bench.
Success uses 25 tasks; resource ratios use 10 shared tasks for SU
and 23 for TRC+SU, relative to self-summarization.
Bars show 95\% task-bootstrap intervals.}
\label{fig:app-summarizer}
\end{figure}

We examine whether changing the summarizer affects task success
and execution efficiency while keeping the agent fixed.
Figure~\ref{fig:app-summarizer} compares self-summarization with
Qwen3.5-9B and Gemma-4-12B for SU and TRC+SU on the 25-task
SWE-bench subset.

SU is more sensitive to the summarizer model: resolve rate falls
from 53.3\% with self-summarization to 37.3\% with Qwen3.5-9B
and 28.0\% with Gemma-4-12B. TRC+SU remains closer to its
68.0\% baseline, achieving 62.7\% and 66.7\%, respectively.
Neither alternative reduces mean end-to-end latency on the
shared eligible tasks. These results suggest that smaller
summarizers do not automatically improve efficiency, and their
effect on task success depends on the compression policy.

\subsubsection{Run outcomes and resource limits}
\label{app:limits}
\begin{figure}[t]
\centering
\includegraphics[width=\linewidth]{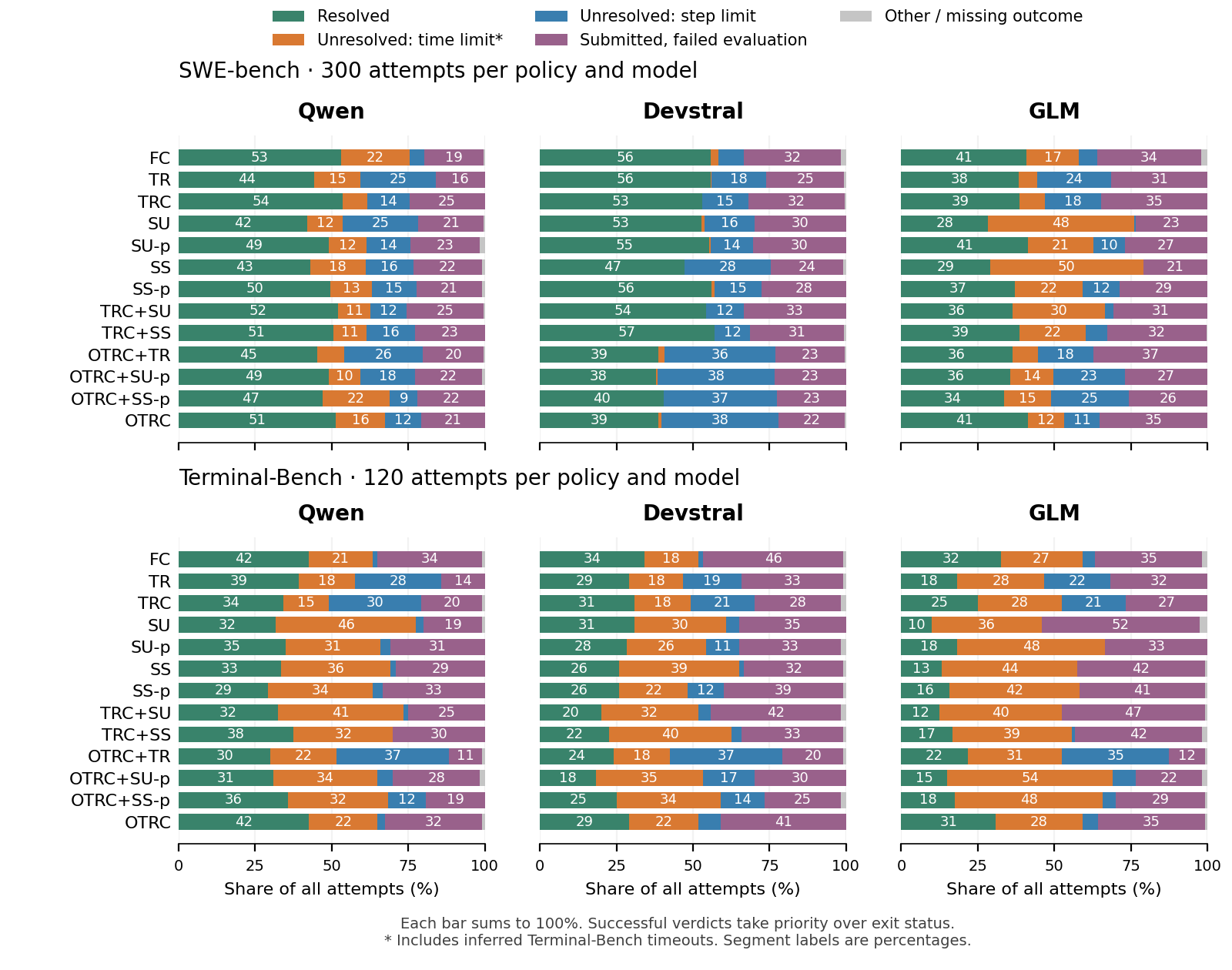}
\caption{Run outcomes at the primary configurations.
Each bar sums to 100\% of attempts, with successful verdicts
taking priority over exit status. Time-limit categories include
inferred Terminal-Bench timeouts; labels show percentages.}
\label{fig:app-limits}
\end{figure}

Figure~\ref{fig:app-limits} distinguishes successful execution
from different unresolved outcomes at the primary configurations.
Each bar includes all three attempts per task: 300 on SWE-bench
and 120 on Terminal-Bench. Successful verdicts take priority;
remaining attempts are classified as time-limited, step-limited,
submitted but unsuccessful, or other/missing outcomes.
The breakdown reveals different failure patterns. On SWE-bench,
GLM's SU and SS runs frequently reach the time limit, whereas
Devstral's step-triggered policies frequently reach the step limit.
On Terminal-Bench, time limits account for a substantial share
of unresolved attempts across models. These patterns provide
context for resource comparisons that exclude time-capped runs.

\subsubsection{Changes in execution length}
\label{app:execution-length}
\begin{figure}[t]
\centering
\includegraphics[width=\linewidth]{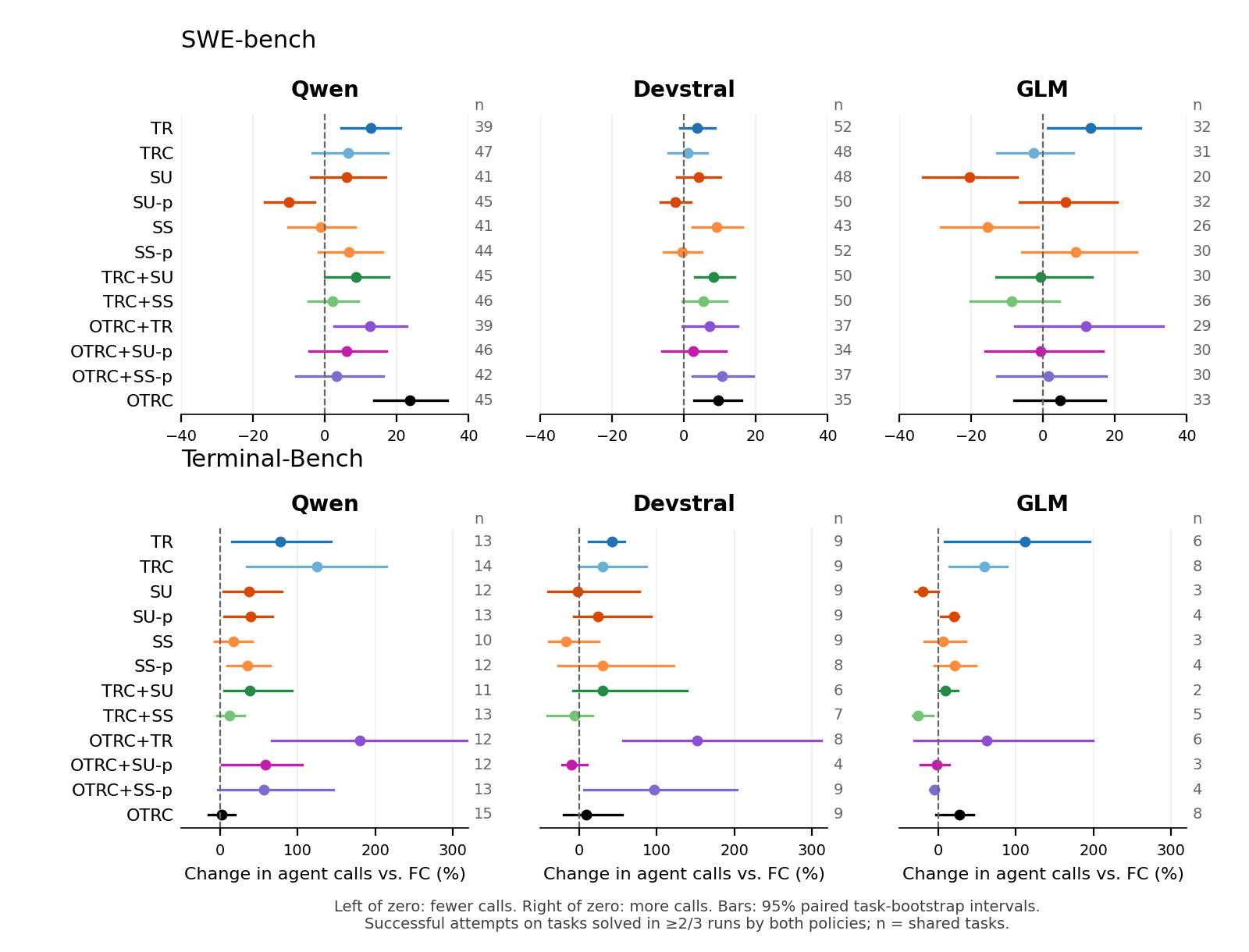}
\caption{Change in agent calls relative to FC on shared solved
tasks. Positive values indicate more calls; bars show 95\%
paired task-bootstrap intervals, and $n$ denotes shared tasks.
Axis scales differ between benchmarks.}
\label{fig:app-steps}
\end{figure}

Figure~\ref{fig:app-steps} compares agent call counts with full
context on tasks solved in at least two of three attempts by
both policies. We average calls over successful attempts within
each task and report the percentage change relative to FC.
Summarizer calls are excluded.

Changes are generally smaller on SWE-bench than on Terminal-Bench.
On Terminal-Bench, OTRC+TR increases mean calls by approximately
180\% for Qwen and 152\% for Devstral. Other policies show smaller
increases or decreases, indicating that compression does not
uniformly lengthen execution. These comparisons describe shared
solved tasks rather than the full cohort; Terminal-Bench estimates
use only 2--15 tasks and should be interpreted alongside their
uncertainty intervals.

